\documentclass{article}
\usepackage{graphicx} 
\usepackage{tablefootnote}
\usepackage[colorlinks=true, allcolors=blue]{hyperref}
\usepackage{authblk, color, bm, bbm, graphicx, epstopdf}
\usepackage{amsmath}
\usepackage{amsfonts}
\usepackage{amssymb}
\usepackage{doi}
\usepackage{ulem}
\usepackage{algorithm}
\usepackage{algpseudocode}
\usepackage{subcaption}
\usepackage{titlesec}
\usepackage{booktabs}
\usepackage{multirow}
\usepackage{array}
\usepackage[letterpaper,top=2cm,bottom=2cm,left=3cm,right=3cm,marginparwidth=1.75cm]{geometry}
\usepackage[numbers,compress,sort]{natbib} 

\usepackage[usenames, dvipsnames]{xcolor}
\usepackage{textcomp}
\usepackage{enumitem}
\usepackage{units}

\usepackage{bm} 

\usepackage{tikz}
\usetikzlibrary{positioning}
\usetikzlibrary{calc}
\usetikzlibrary{shapes.geometric}
\usetikzlibrary{topaths}
\usetikzlibrary{external}
\usetikzlibrary{fadings,decorations.pathreplacing} 

\usepackage{pgfplots}
\usetikzlibrary{shapes,decorations.pathmorphing,patterns}
\pgfplotsset{compat=newest}
\usepgfplotslibrary{fillbetween}

\graphicspath{{figures/}}

\usepackage[colorinlistoftodos,prependcaption]{todonotes}

\title{An Online Machine Learning Multi-resolution Optimization Framework for Energy System Design Limit of Performance Analysis}

\date{2026}
\author[1]{Oluwamayowa O. Amusat\thanks{Authors contributed equally to this work.}}
\author[*2]{Luka Grbcic  \thanks{Corresponding author}} 
\author[3]{Remi Patureau\thanks{Work done while employed at Lawrence Berkeley National Laboratory}}
\author[4]{M. Jibran S. Zuberi}
\author[1]{Dan Gunter}
\author[3]{Michael Wetter}
\affil[1]{\small Scientific Data Division, Computing Sciences, Lawrence Berkeley National Laboratory (LBNL), Berkeley, CA 94720}
\affil[2]{\small Applied Mathematics \& Computational Research Division (AMCR), Computing Sciences, Lawrence Berkeley National Laboratory (LBNL), Berkeley, CA 94720}
\affil[3]{\small Building \& Industrial Energy Systems (BIES) Division, Energy Technologies Area, Lawrence Berkeley National Laboratory (LBNL), Berkeley, CA 94720}
\affil[4]{\small Energy Analysis Division, Energy Technologies Area, Lawrence Berkeley National Laboratory (LBNL), Berkeley, CA 94720}

\begin{document}

\maketitle

\begin{abstract}
Designing reliable integrated energy systems for industrial processes requires optimization and verification models across multiple fidelities, from architecture-level sizing to high-fidelity dynamic operation. However, model mismatch across fidelities obscures the sources of performance loss and complicates the quantification of architecture-to-operation performance gaps. We propose an online, machine-learning–accelerated multi-resolution optimization framework that estimates an architecture-specific upper bound on achievable performance while minimizing expensive high-fidelity model evaluations. We demonstrate the approach on a pilot energy system supplying a 1 MW industrial heat load. First, we solve a multi-objective architecture optimization to select the system configuration and component capacities. We then develop an machine learning (ML)-accelerated multi-resolution, receding-horizon optimal control strategy that approaches the achievable-performance bound for the specified architecture, given the additional controls and dynamics not captured by the architectural optimization model. The ML-guided controller adaptively schedules the optimization resolution based on predictive uncertainty and warm-starts high-fidelity solves using elite low-fidelity solutions. Our results on the pilot case study show that the proposed multi-resolution strategy reduces the architecture-to-operation performance gap by up to 42\% relative to a rule-based controller, while reducing required high-fidelity model evaluations by 34\% relative to the same multi-fidelity approach without ML guidance, enabling faster and more reliable design verification. Together, these gains make high-fidelity verification tractable, providing a practical upper bound on achievable
operational performance
prior to industrial deployment.
\end{abstract}
\vspace{5pt}
\newpage
\section{Introduction}
Integration across mechanical and electrical systems is a key enabler
to achieve cost-effective, grid-flexible energy systems as such integration
allows optimizing the size of equipment for storing and converting energy
across energy carriers under consideration of dynamic energy rates.
For example, for given utility rates and load profiles, one could
trade off investments in batteries and thermal storage tanks
for energy storage, and heat pumps and gas boilers for heat production.
The overall techno-economic performance depends on
both how equipment is sized and how it is operated, requiring a
co-optimization across electrical, mechanical, and controls domains.

There are two general approaches to this co-optimization challenge.
Either one uses simplified process and control models to formulate
an optimization problem that can be solved efficiently and robustly, such
as a linear programming problem or a nonlinear programming problem. 
These types of models assume
perfect knowledge of the system and the future,
and ignore fast transients and control switches that would
lead to non-differentiable functions.
We will call these types of models \emph{optimization models}.
Alternatively, one may build detailed process and control models,
such as by formulating a differential algebraic system of equations
with continuous-time and discrete event dynamics.
Such models serve well for verification of a design, but determining
equipment sizes and control schedules can only be optimized heuristically,
and their simulation can be computationally expensive, sometimes making even
heuristic optimizations impractical.
We will call these models \emph{verification models}.

For complex energy systems such as a power plant that provides heat for an industrial process
or cooling for a building, district, or data center, there
is no model that can serve as both optimization and for verification~\cite{WetterSulzer2024}.
Platform-Based Design (PBD) is a process that formalizes how to transition between
models of  different fidelity, allowing holistic design space exploration
and de-risking implementation of integrated systems~\cite{Sangiovanni2007:1,SulzerWetterMutschlerSangiovanni2023}.
However, as the optimization model and the verification model are different,
they have an inherent model mismatch and corresponding performance gap.
The different models will have different
approaches and assumptions, such as
perfect knowledge of the process and the future in optimization models
compared to
only measurable quantities and imperfect predictions 
with closed loop stability constraints and operational constraints
for verification models.
As one progresses in the design from the optimization to the implementation with
verification, it is not clear what part of the performance gap is due to mismatch of the process models and due to implementation of the control.
However, knowing the performance gap due to control implementation is important,
as such knowledge allows designers to judge whether the control
is near-optimal, and further attempts to improving it would be fruitless.


Addressing this knowledge gap is the core contribution of the paper. One potential strategy involves a direct brute-force optimization of the verification model. 
However, the optimal control problem is a constrained, non-differentiable black-box of high dimensionality, which makes the problem computationally expensive and achieving a competitive result difficult at best.
An exhaustive search is typically required to close the performance gap with optimal solutions. To overcome these challenges, we propose an integrated framework that leverages multi-resolution optimal control and machine learning (ML) for further acceleration. This approach aims to improve both the tractability and computational efficiency when solving limit of performance analysis problems. 

Multi-resolution optimal control approaches have been shown to be beneficial for improving the tractability of general optimal control problems in power and energy storage systems \cite{abdulla2017multi, jain2008trajectory, zhang2023multi, zou2018wavelet}. The fundamental principle of multi-resolution optimal control is to leverage optimal control sequences from a coarse-grained discretization to inform the optimization process at a finer-grained level \cite{Pol97:1, jain2008trajectory}. 
This strategy is designed to converge towards a solution with fewer evaluations of
the computationally expensive function.

Similarly, ML techniques have been shown to be able to accelerate the solution of optimal control in energy systems problems \cite{perera2019machine, bre2020efficient, prina2024machine, stoffel2023safe,ratz2024identifying}. More specifically, optimal control in energy systems heavily relies on ML methods such as Deep Reinforcement Learning (DRL) for rapid, policy-based decision-making under uncertainty \cite{li2023deep, michailidis2025reinforcement}. Furthermore, ML techniques enhance classical control frameworks, such as utilizing algorithms based on Koopman theory to establish accurate, simplified models for Model Predictive Control (MPC) in complex components like Pumped Storage Systems \cite{cai2021machine}, while hybrid solutions like Reinforced Predictive Control (RL-MPC) are emerging to ensure constraint satisfaction and continuous learning in building energy management \cite{arroyo2022reinforced}. For energy system design problems, ML surrogate models are employed to emulate high-fidelity energy system simulation tools, serving as fast, reliable replacements for costly iterative calculations \cite{rulff2025systematic, ledee2025improved}. To improve the efficiency of black-box optimization, ML surrogate models are deployed in two primary ways: either to provide a good initial guess to warm start the main optimization algorithm \cite{grbcic2024efficient, favaro2024multi}, or as a computationally efficient replacement for the original black-box function, thereby conserving computational resources \cite{liang2024survey, liu2023surrogate}. 

Furthermore, in the field of online trained ML models applied to energy system problems, \citet{stoffel2023safe} introduce a methodology that integrates online learning and novelty detection in data-driven MPC to enhance the reliability and safety of building energy systems by continuously retraining models and detecting extrapolation events to switch to robust fallback controllers when necessary. \citet{gao2025online} present an online optimization method for integrated energy systems that combines deep learning for accurate forecasting of wind and photovoltaic power generation with an accelerated optimization algorithm. \citet{sha2025online} introduce a data-driven MPC strategy that employs an online learning-enhanced encoder-decoder ML model to optimize HVAC performance in residential buildings.

While existing research has separately applied multi-resolution methods to improve tractability and ML techniques to accelerate optimization, these approaches have not been applied to limit of performance problems in energy system design, or synergistically integrated into a unified framework capable of solving high-dimensional, black-box optimal control problems efficiently. Our work makes three main contributions to the state-of-the-art in energy system and controls co-optimization, and in limit of performance analysis:
\begin{itemize}
\item We introduce a novel, ML-accelerated, multi-resolution framework for efficient numerical approximation of the limit of performance in energy systems whose transient evolution is specified in differential-algebraic system of equations whose state trajectories are not differentiable in the control schedule. This integrated approach improves tractability over direct optimization and reduces computational cost by intelligently replacing an expensive, long-horizon optimization stage with an online-trained ML surrogate. As demonstrated in Section \ref{subsec:lop-analysis}, when applied specifically to our case study, this framework establishes a performance benchmark that reveals a potential for over 10\% in operational cost savings compared to a standard rule-based controller.
\item We demonstrate, through an ablation study, the synergistic effect of the framework's core components. We show that both the multi-resolution structure (which provides strategic guidance) and the elite solution seeding (which preserves high-quality prior solutions) are independently critical for achieving robust convergence and computational efficiency.
\item We present a novel supervisory control logic that leverages the ML model's uncertainty quantification to intelligently balance computational cost and solution quality. This uncertainty-aware mechanism ensures both computational efficiency, by using the fast ML model when it is confident, and robustness, by triggering a full optimization to explore and gather new data when uncertainty is high or conditions change.
\end{itemize}

\begin{figure}[h]
  \centering
  \includegraphics[width=\textwidth]{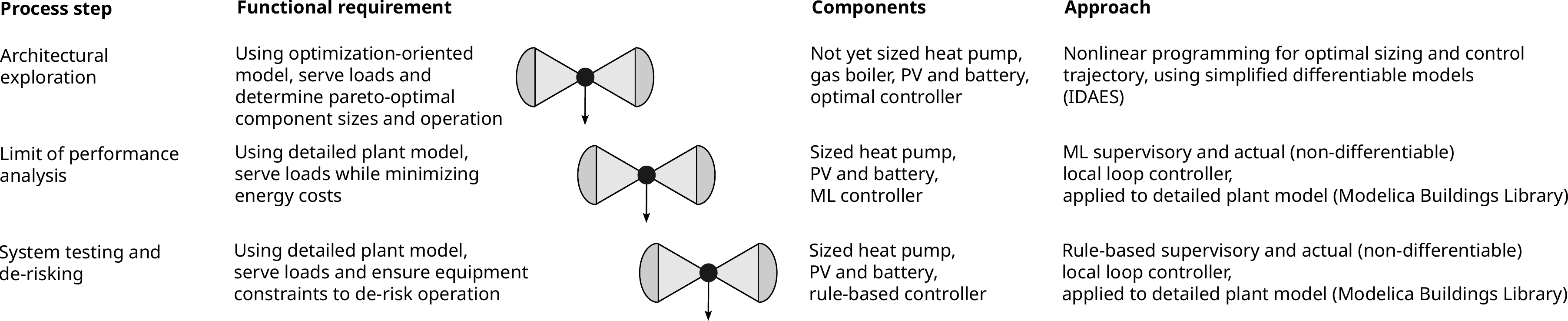}
  \caption{Three platform layers used in the Platform-Based Design of the industrial plant and its controller.}
  \label{fig:pbd}
\end{figure}

We demonstrate our approach to the limit of performance analysis in energy system and controls co-design of an industrial heating process, for which we design the system
and its implementable control using PBD, consisting of three layers of abstraction as shown in Figure~\ref{fig:pbd}:
In the first platform layer, we co-optimize electrical and mechanical system together with optimal
operating schedules using nonlinear programming. This gives specification for
equipment sizes and insight into how to operate the system.
For the second and third platform layer, 
we implement a verification model that is partitioned into a process model,
a supervisory controller and local loop controllers.
In the second platform layer, we use the process and the local loop control models together with an ML-based supervisory controller to compute a numerical  approximation of limit of performance for the system, thereby providing an upper bound for the performance of the verification model under the assumption that the limit of performance analysis found the global optimal control sequence.
In the third platform layer, we replace the ML-based controller with a rule-based supervisory controller,
and compare its performance against the limit of performance computed using the ML-based
supervisory controller applied to the same model of the process and the local loop controllers.
The rule-based controllers ensure that practical constraints such as equipment lockout and operating envelopes are maintained.
We note that while we used a rule-based controller,
the limit of performance analysis does on depend on that choice;
the same process could have been done using for example a robust model predictive controller
instead of the rule-based controller in which case one would see how much performance has
been left on the table due to provision of robustness guarantees.

The computation of this limit of performance is achieved through our novel ML-accelerated, multi-resolution optimal control framework. The foundation of the framework is a multi-resolution strategy, where an initial, coarse-grained "exploratory" optimization over a long horizon determines a strategic terminal target value (e.g. battery State of Charge), which then guides subsequent, finer-grained optimization stages. Our core contribution is to accelerate this process by training an online ML model with historical forecast data to predict this terminal target value based on current forecasts. Crucially, the ML model also provides an estimate of its own prediction uncertainty, which is central to the algorithm's performance. 

The rest of the paper is structured as follows: Section \ref{sec:pilot-case-description} presents a description of the industrial heating system that will be the focus of this work. Section \ref{sec:arch-exploration} presents at a high level the nonlinear programming optimization formulation for the architectural exploration and system sizing problem. Next, we briefly describe the verification model, followed by a detailed description of our proposed ML-Accelerated optimal control framework (Section \ref{sec: energy-system-control}). Section \ref{sec: results} presents a detailed analysis of our results for the case study, covering both the architectural optimization and optimal control problems. This is followed by a brief discussion of the limitations of the proposed approach, and potential future research directions to expand this work (Section \ref{sec: limitations}).  



           
\section{Pilot System Description}\label{sec:pilot-case-description} 

\begin{figure}[h]
  \centering
  \includegraphics[width=0.9\textwidth]{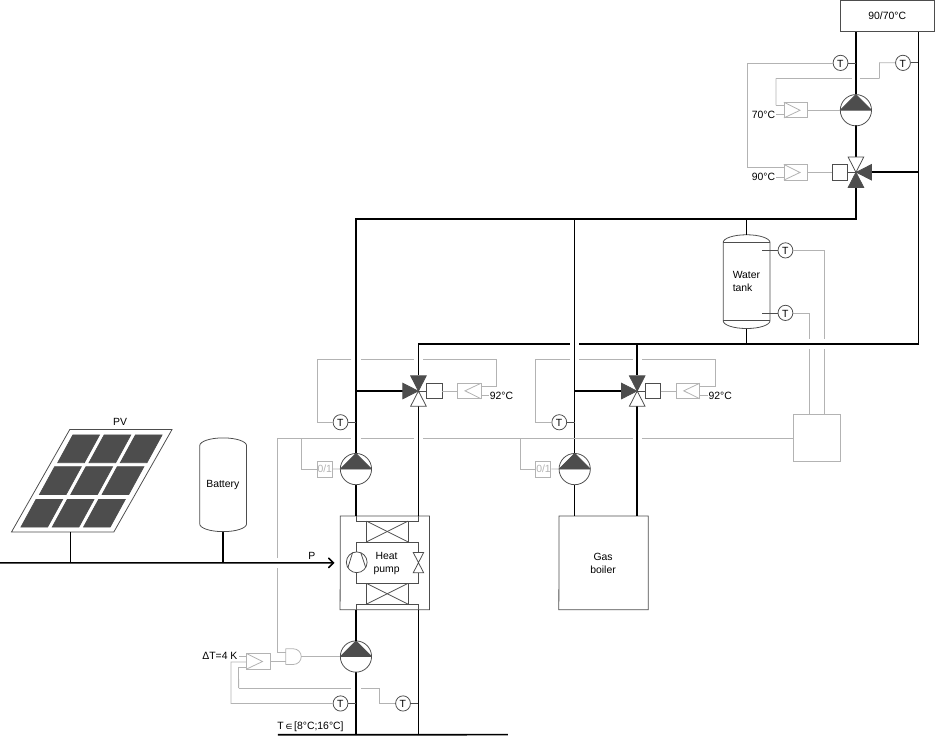}
  \caption{Schematic diagram of the energy system.}
  \label{fig:plant}
\end{figure}

In this section, we describe the energy system and the control intent that is used
for the subsequent co-optimization of the energy and control system.

Figure~\ref{fig:plant} shows the schematic diagram of the system.
The objective is to serve an industrial heat load $\dot Q_l(\cdot)$ daily from $8$am to $8$pm.
The industrial process is controlled using a heat exchanger
that requires on the plant side to maintain
a supply temperature of $90^\circ \mathrm C$ and a return temperature of $70^\circ \mathrm C$.
The energy and control system should be designed to minimize total annualized cost and
energy use, subject to meeting the load at the specified temperatures.

The energy system shown in Figure~\ref{fig:plant} is as follows:
Hot water can be produced either by a heat pump which uses as its source temperature
a district energy system whose temperature varies between
$8^\circ \mathrm C$ and $16^\circ \mathrm C$, or a gas boiler.
The heat pump has internal control to track the leaving evaporator water outlet temperature,
subject to sufficient capacity, and the heat provided by the gas boiler is modulating.
The heat pump evaporator loop has a variable speed pump to track an evaporator
temperature difference of $\Delta T=4 \, \mathrm K$.
On the heat supply side, both heat production units have on/off pumps and three-way valves
that regulate the leaving temperature to at least $92^\circ \mathrm C$. The heat can either
be directly used by the load or stored in a stratified water tank.
The circulation pump at the load is a variable speed pump, controlled to track the required return temperature.
Electricity can be consumed from or fed back to the electrical grid.
The plant has a PV array and a battery that allows storing electricity for the heat pump.\footnote{Compared to the required heat energy, the energy of the circulation pumps is negligible and hence not accounted for in the optimization.}

To protect equipment from short-cycling and associated wear-out, we enforce the following lockout constraints based on engineering judgment:
If the heat pump is switched on, it must operate for at least $30$ minutes,
and when it is switched off, it must remain off for at least $10$ minutes.
If the gas boiler is switched on, it must remain on for at least $15$ minutes,
but there is no minimum lockout time.

For the solar irradiation, TMY3 weather data~\cite{WilcoxMarion2008} for the Sacramento international airport are used.
\section{Architectural Exploration via Nonlinear Programming}\label{sec:arch-exploration}

For the system described above, we determine the optimal architecture and component sizes for the energy system by solving a nonlinear programming (NLP) optimization problem. Given input data on the thermal load demand, solar irradiation, and electricity and gas costs, we solve a multi-objective optimization problem to minimize the total annualized cost $TAC(\mathbf x)$ and total energy imported $TE(\mathbf x)$, 
\begin{equation}
\begin{aligned}
\min_{\mathbf{x}} \quad 
    \mathbf{f}(\mathbf{x})
    &= 
    \begin{bmatrix}
        f_1(\mathbf{x}) = \mathrm{TAC}(\mathbf{x}) 
            = C_{\mathrm{cap}}(\mathbf{x}) \cdot \mathrm{CRF} + C_{\mathrm{op}}(\mathbf{x}) \\[2mm]
        f_2(\mathbf{x}) = \mathrm{TE}(\mathbf{x}) 
            = \displaystyle 
            \int_{t_0}^{t_f} E_{\mathrm{grid}}(t, \mathbf{x}) \, dt
            + \int_{t_0}^{t_f} Q_{\mathrm{boiler}}(t, \mathbf{x}) \, dt
    \end{bmatrix}, \\[1mm]
\text{s.t.} \quad 
    & g_i(\mathbf{x}) \le 0, \quad i \in \{1, \dots, m\},\\
    & h_j(\mathbf{x}) = 0, \quad j \in \{1, \dots, z\},
\end{aligned}
\end{equation}

where $\mathbf{x}$ is the vector of decision variables representing the capacities of the individual units and the variables defining the operation of each equipment at each point in time; $C_{\mathrm{cap}}(\mathbf{x})$ and $C_{\mathrm{op}}(\mathbf{x})$ are the capital and operating costs, respectively; $\mathrm{CRF}$ is the capital recovery factor; $E_{\mathrm{grid}}(t, \mathbf{x})$ is the energy purchased from the grid at time $t$; and $Q_{\mathrm{boiler}}(t, \mathbf{x})$ is the energy generated from the natural gas imported for the boiler at time $t$. Equality constraints $h_j(\mathbf{x})$ on the optimization problem include constraints enforcing the mass and energy balances within the system, performance equations, and  mechanistic models of the main energy system  components (PV, gas boiler, heat pump, battery, storage tank, load).  Inequality constraints $g_j(\mathbf{x})$ include the thermal load satisfaction and other operational constraints, as well as bounds on the generation and storage capacities of the different system components. Details about the process and costing models for the individual components are presented in {Appendix \ref{app:process-models}. 

The architectural optimization model is implemented and solved using the IDAES Integrated Platform~\cite{lee2021}, an algebraic modeling framework with advanced mathematical tools for solving large scale optimization problems within the Python ecosystem. We adopt the multi-period modeling approach~\cite{rawlings2022multiperiod, rao2024}, which converts a dynamic optimization problem into a series of discrete models linked together over a time horizon with rate-limiting constraints. 
This is accomplished by approximating the dynamic behavior of the system components by algebraic equations via a forward-difference discretization scheme, enabling implementation in IDAES with an hourly time horizon.
With this multi-period modeling approach, each time step in the horizon is implemented as a separate algebraic model, and the different models are connected together by user-defined set of variables and constraints and solved simultaneously. For all generation and storage units $k$, we implement linking constraints that force the installed capacities $S$ of the units to be the same across all time periods,
\begin{equation}
S_{k,t} = S_{k,t-1}, \quad t \in\{ 1, 2, \dots, N\}.
\end{equation}

For the storage units (battery, tank), linking constraints also connect the final state of charge of the previous time horizon to the starting state of charge of the current time horizon, 
\begin{equation}
E_{k,t,start} = E_{k,t-1,end}, \quad t \in\{ 1, 2, \dots, N\}; k\in\{{\text{{battery, storage tank}}}\}.
\end{equation}

The architectural optimization problem is formulated as a boundary value problem, with constraints that enforce that the storage levels for the battery and heat tank at the start and end of operation are equal,
\begin{equation}
E_{k,1,start} = E_{k,N,end}, \quad k\in\{{\text{{battery, storage tank}}}\}.
\end{equation}

We solve the multi-objective problem with the $\epsilon$-constraint method, converting the single multi-objective problem into a series of single-objective cost minimization problems in $\epsilon$, 
\begin{equation}
\begin{aligned}
\min_{\mathbf{x}} \quad & \mathrm{TAC}(\mathbf{x}) \\[2mm]
\text{s.t.} \quad 
    & \mathrm{TE}(\mathbf{x}) \le \epsilon, \\ 
    & g_i(\mathbf{x}) \le 0, \quad i \in \{1, \dots, m\}\\
    & h_j(\mathbf{x}) = 0, \quad j \in \{1, \dots, z\}.
\end{aligned}
\end{equation}

with the range of $\epsilon$ $(TE_{LB} \le \epsilon \le TE_{LB})$ determined by solving two single objective problems minimizing the total annualized cost $\mathrm{TE}_{UB}$ and total imported energy $\mathrm{TE}_{LB}$, respectively.

We solved the NLP architectural optimization problem with IPOPT~\cite{wachter2006}. The discretized nonlinear system of algebraic equations for the full year ($t=1, 2,\ldots,8760 \text{h}$) consists of over 1.4M equations and constraints, with 35,040 degrees of freedom (four per time step).

The solution of the architectural optimization problem is a set of Pareto-optimal points that are a trade-off between TAC and imported energy from the grid. As part of the optimization solution, the model also returns an idealized operating strategy for each design that minimizes the operating costs. However, the architectural optimization problem assumes 
perfect knowledge of the future (i.e., solar irradiation and utility rates are known for the whole year a priori), does not capture full dynamics such as startup, shutdown, lockouts, and fast transients, and ignores all non-differentiable control implementations such as actuator saturation and switches between mode of operations. Thus, while the optimal control profiles obtained at the architectural optimization level will capture the slow dynamics of energy storage well, they can not be applied to verify operation and controls, and de-risking is necessary using verification models that capture all these intricacies. However, the idealized control trajectories enable us to optimally size equipment under consideration of the dynamic system operation, and they provide a valuable insights for operating the systems.



\section{Energy System Control}\label{sec: energy-system-control}
The architectural optimization process described in the previous section determines the optimal unit capacities for the different components of the energy system. Given those capacities, the next task is to design control logic that gets the best possible performance out of the specified architecture, given the additional controls and dynamics not captured by the optimization model. For this, we use a verification model of the system implemented in Modelica, an object-oriented, equation-based modeling language~\cite{MattssonElmqvist1997:1}. The verification model effectively represents a digital representation of the actual process. In this section, first, we describe the Modelica model. We follow that with a description of a rule-based controller, which provides a controller as is typical in such thermal systems. Finally, we provide details of our proposed ML-accelerated optimal control strategy which is used to approximate the limit of performance. This limit of performance allows us to quantify how much performance is lost due to the
rule-based controller relative to the highest performing controller that we were able to develop, but which may not be sufficiently robust for an actual implementation.

\subsection{Modelica model}

Two models were developed in Modelica using the Modelica Buildings Library~\cite{WetterZuoNouiduiPang2014}.
Modelica is an equation-based, object-oriented language~\cite{MattssonEtAl1999} which can be used to simulate models that couple thermal systems, fluid flow systems, electrical systems and control.
Modelica can properly resolve the continuous time dynamics, discrete time dynamics and event-driven dynamics that is needed for verification of such coupled physical and control systems.
In Modelica, components are encapsulated models that
expose interface variables, such as mass flow rate, pressure
and enthalpy of a pipe, through ports, and through these
ports, components can be connected to assemble a system.
For our application, the Modelica model results in a differential algebraic system of equations
that is governed by continuous dynamics as well as time- and state-events.
Thus, the resulting system model is not differentiable in equipment sizes or in control inputs.
To translate Modelica models to executable code, we used Dymola 2026 on Linux Ubuntu.

The two models differ as follows:
One model is a complete model of the plant with rule-based local loop control,
but with supervisory control set points exposed
as inputs, allowing the limit-of-performance calculation
to vary the supervisory control in order to maximize performance.
The exposed control inputs were
charging and discharging rates of the battery
and the mass flow rate of the heat pump condenser. By controlling the mass flow rate,
the limit-of-performance controller can dispatch the equipment.
This model is used in the 2nd platform layer shown in Figure~\ref{fig:pbd}.
The other model adds rule-based supervisory controls,
described in Section~\ref{sec:rule_based_control},
and is used in the 3rd platform layer.
Table~\ref{tab:Modelica-model-components} summarizes the key Modelica modeling components used in our analysis; further details may be found in Appendix \ref{app:modelica-models}.

\begin{table}
\caption{Modelica model components \label{tab:Modelica-model-components}}
\centering{}%
\begin{tabular}{cl}
\toprule 
Component & Modelica model\tabularnewline
\midrule 
Heat pump & \textit{Buildings.Fluid.HeatPumps.Carnot\_TCon}\tablefootnote{\url{https://simulationresearch.lbl.gov/modelica/releases/v12.0.0/help/Buildings_Fluid_HeatPumps.html}}\tabularnewline
Gas boiler & \textit{Buildings.Fluid.Boilers.BoilerPolynomial}\tablefootnote{\url{https://simulationresearch.lbl.gov/modelica/releases/v12.0.0/help/Buildings_Fluid_Boilers.html}}\tabularnewline
Thermal storage & \textit{Buildings.Fluid.Storage.StratifiedEnhanced}\tablefootnote{\url{https://simulationresearch.lbl.gov/modelica/releases/v12.0.0/help/Buildings_Fluid_Storage.html}}\tabularnewline
Battery & \textit{Buildings.Electrical.DC.Storage.Battery}\tablefootnote{\url{https://simulationresearch.lbl.gov/modelica/releases/v12.0.0/help/Buildings_Electrical_DC_Storage.html}}\tabularnewline
PV & \textit{Buildings.Electrical.DC.Sources.PVSimple}\tablefootnote{\url{https://simulationresearch.lbl.gov/modelica/releases/v12.0.0/help/Buildings_Electrical_DC_Sources.html}}\tabularnewline
\bottomrule
\end{tabular}
\end{table}


\subsection{Rule-based control}
\label{sec:rule_based_control}

The rule-based controller attempts to minimize the operating
cost by dispatching the heat pump or boiler, and charging
and discharging the battery, based on the
time-varying electricity price.

We implemented a rule-based strategy as follows:

When the process load is active (from 8am to 8pm), the heat pump and the gas boiler are commanded on.
When the process load is inactive, the heat pump charges the hot water tank during the hours when the electricity cost is the least expensive, provided the electricity cost is
    below $C_{ele} < 0.13$ \$/kWh. This limit is computed based
    on the comparison of gas versus electricity costs, taking into
    account the gas boiler efficiency and the coefficient of
    peformance of the heat pump as
\begin{equation} \label{eq:comp_cost}
	C_{ele} <  C_{gas} \, \frac{COP}{\eta_{boiler}}.
\end{equation}

The battery operation is based on the hourly electricity prices.
When the process load is active (from 8am to 8pm), the battery is discharging during the 4 hours when the electricity is the most expensive.
The battery is charging during the 4 hours when the electricity is the least expensive. 

\begin{figure}
	\centering
	\includegraphics[width=0.9\linewidth]{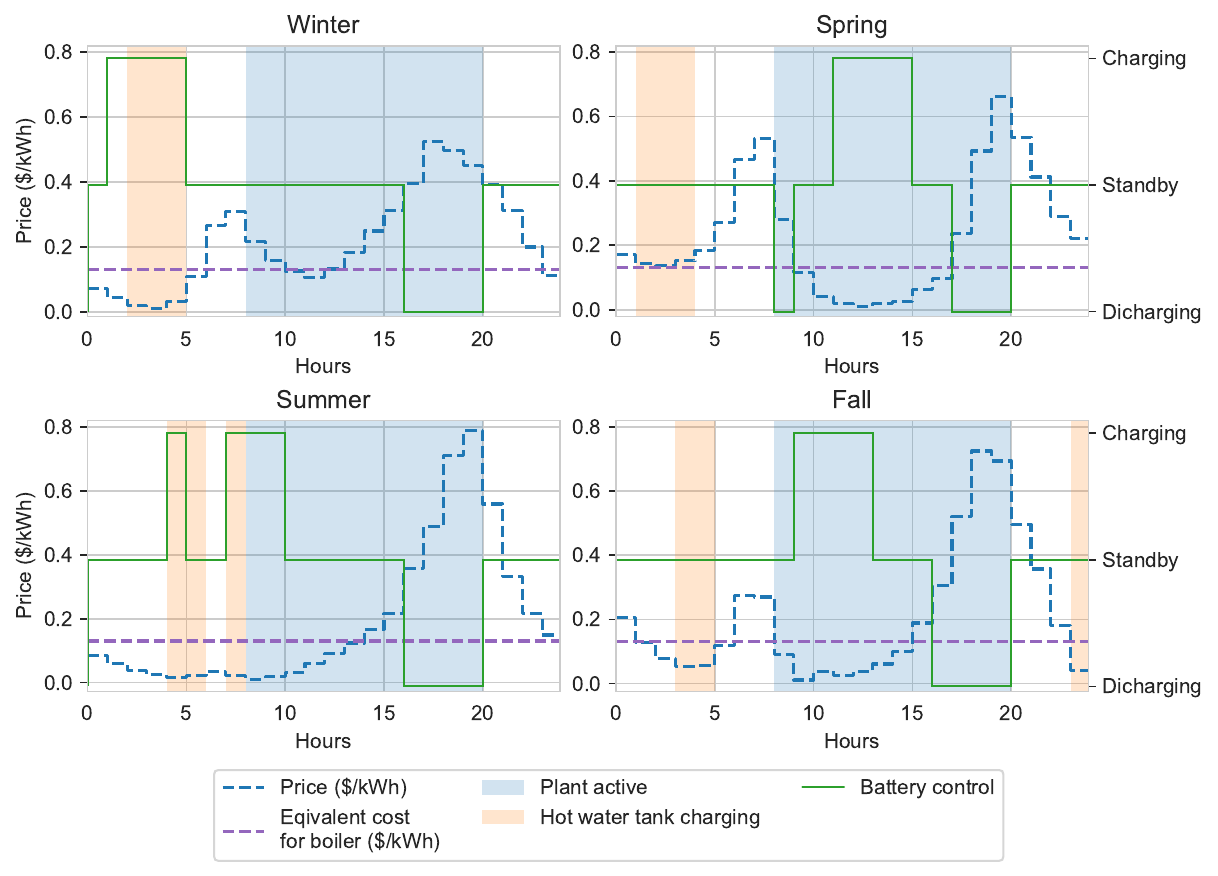}
	\caption{Charging and discharging of the battery (green), and hot water tank charging (orange). The electricity price and the equivalent for gas are shown with dotted lines.}
	\label{fig:season_price_rule}
\end{figure}
Figure~\ref{fig:season_price_rule} shows the resulting
control strategy.

\subsection{ML-Accelerated Optimal Control}

This section details a general framework for accelerating a multi-resolution, receding-horizon optimal control problem using a ML model. We first present the generalized mathematical formulation of the optimal control problem. We then describe the ML-accelerated algorithm, which adaptively decides when to leverage a predictive model versus performing a full, computationally expensive optimization. Finally, we provide the specific implementation details and hyperparameters of the framework. 

\subsubsection{General Optimal Control Formulation}

We consider a discrete-time optimal control problem solved in a receding-horizon fashion. The system's evolution is defined over a finite time horizon, which is discretized into $H$ steps of duration $\Delta t$.

The state of the system at time step $k$ is represented by a vector $x_k \in \mathbb{R}^n$, and the control inputs are given by $u_k \in \mathcal{U} \subseteq \mathbb{R}^p$. These control inputs are constrained by element-wise lower and upper bounds, represented by vectors $u_{\min}$ and $u_{\max}$, ensuring that the inputs remain within their limits.

The full decision vector for the optimization horizon is the concatenation of all control inputs, $U = [u_0^\top, u_1^\top, \dots, u_{H-1}^\top]^\top \in \mathcal{U}^{m}$, where $m=pH$.

The objective is to find the sequence of control inputs $U$ that minimizes a total cost function, $J(U)$, subject to the system's dynamics and operational constraints. The general optimal control problem is formulated as
\begin{equation}
\begin{aligned}
\min_{U \in \mathbb{R}^{m}} \quad & J(U), \\
\text{s.t.} \quad &
x_{k+1} = f(x_k, u_k, w_k), \quad \text{for } k \in \{0, \dots, H-1\}, \\
& x_k \in \mathcal{X}, \\
& u_k \in \mathcal{U}, \\
& x_0 \text{ is the known initial state.}
\end{aligned}
\end{equation}
Here, the function $f(\cdot, \cdot, \cdot)$ is a discrete-time, non-differentiable function that describes the system dynamics. The vector $w_k$ represents known exogenous forecasts, such as weather or market prices, that influence the system's evolution. The sets $\mathcal{X}$ and $\mathcal{U}$ represent the feasible states and controls, respectively, enforcing the physical operational limits of the system, including applicable box bounds.

The cost function $J(U)$ is defined as the sum of a stage cost $\ell(x_k, u_k, w_k)$, which is summed over the horizon, and a terminal cost $V_f(x_H)$, which is evaluated at the final state $x_H$ as
\begin{equation}
J(U) = \sum_{k=0}^{H - 1} \ell(x_k, u_k, w_k) + V_f(x_H).
\end{equation}

\subsubsection{ML-Accelerated Multi-Resolution Strategy}

Solving the above optimization problem can be computationally intractable, especially when the dynamics function $f(\cdot,\cdot,\cdot)$ is expensive to evaluate, when it is not continuously differentiable, as is the case for our problem because
of local controllers, or when the horizon $H$ is long and requires exhaustive search. To address this, we employ a multi-resolution strategy composed of three distinct stages: a long-horizon exploratory stage ($r=\hat{e}$), a low-resolution stage ($r=\hat{l}$), and a high-resolution stage ($r=\hat{h}$). The problem is first solved at the exploratory resolution, which uses a coarse time discretization, i.e., a large $\Delta t$, to cover a long total time duration. This allows the optimizer to find a good long-term strategy while keeping the number of decision steps $H_{\hat{e}}$, and thus the problem’s dimensionality, computationally tractable. Subsequently, this strategy is refined in two stages with progressively finer detail. The low-resolution stage ($r=\hat{l}$) uses a smaller time step ($\Delta t_{\hat{l}} < \Delta t_{\hat{e}}$) over a shorter total duration, resulting in a new planning horizon $H_{\hat{l}}$. Finally, the high-resolution stage ($r=\hat{h}$) uses the finest time step ($\Delta t_{\hat{h}} < \Delta t_{\hat{l}}$) over the duration for a horizon of $H_{\hat{h}}$ that is equal to $H_{\hat{l}}$. For all stages, the number of decision steps $H_r$ is chosen to ensure the problem remains computationally tractable.

Solving the optimization problem at any resolution is computationally demanding, as the system model's non-differentiability necessitates a solver that performs thousands of function evaluations. The core of our proposed method is to use machine learning to further accelerate this overall process by strategically replacing the computationally expensive part: the initial, coarse-grained exploratory optimization. This exploratory stage is typically used to identify a key long-term target, such as a desired terminal state for subsequent, finer-grained optimization stages. The algorithm intelligently decides whether to run this expensive exploratory optimization or to use a fast prediction of the desired terminal state from an ML model. The logic is detailed in Algorithms \ref{alg:main_loop_general}, \ref{alg:determine_target_general}, and \ref{alg:find_control_general}. The flowchart of the algorithm is shown in Figure \ref{fig:algortihm_flowchart}.

\begin{figure}
    \centering
    \includegraphics[width=1\linewidth]{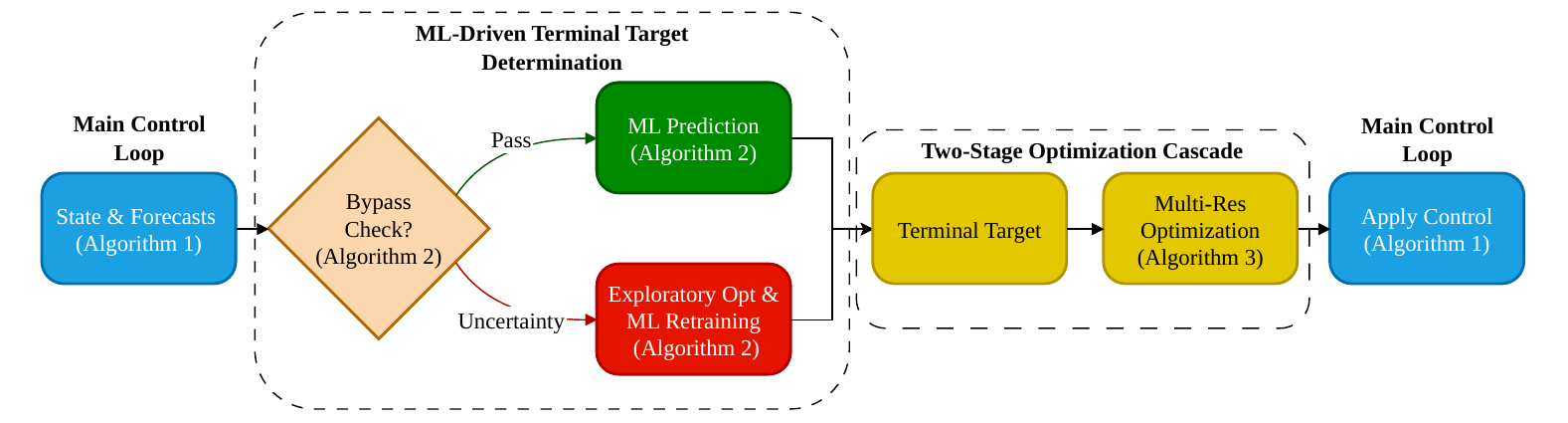}
    \caption{The flowchart of the ML-Accelerated multi-resolution control strategy.}
    \label{fig:algortihm_flowchart}
\end{figure}

\paragraph{Algorithm Notation}
Before presenting the algorithms, we define the key data structures. The elite pool $\mathbf{\mathcal{E}}$ is a cumulative set containing the entire group of final solutions, denoted as the decision vectors $U$, from each preceding high-resolution optimization cycle. The training datasets $\{\mathcal{D}_j\}_{j=1}^{N_v}$ store pairs of input and output features $(\phi_k, v_{\mathrm{target}, j})$ used to train the specific ML model for each terminal target component. The history of applied controls $\mathbf{U}_{\text{hist}}$ is a set of the optimal high-resolution control sequences that were used to advance the system state. In the context of the optimizer, $U^{(r)*}$ refers to the single best solution vector found at resolution $r$, while $\mathcal{P}^{(r)*}$ refers to the entire final set of candidate solutions obtained from the solver, hence $U^{(r)*} \in \mathcal{P}^{(r)*}$.

\paragraph{Algorithm Parameters}
Given the non-differentiable nature of the system dynamics, we employ a population-based evolutionary algorithm to solve the optimal control problem. The optimization process and the ML-driven logic are governed by several key parameters For each resolution $r$, the solver uses a size of $\mathcal{S}_r$ candidate solutions over each iteration. The computational effort is controlled by a black-box function evaluation budget, which has a default value $E_r^{\text{def}}$ and is temporarily increased to a larger initial value $E_r^{\text{init}}$ to encourage broader exploration at the start of the optimal control or after a reset. A reset is triggered by comparing the latest price forecast $\mathbf{p}_k^{\text{reset}}$ against the forecast from the previous cycle $\mathbf{p}_{\text{last}}$, $\mathbf{p}_{\text{last}}$, allowing the controller to detect fundamental shifts in the market regime that invalidate previous optimization results. A cycle constitutes one complete iteration of the control process: solving the optimization problem for a planning horizon $H_{\hat{h}}$ and then applying the initial control actions to the system. The core of our method relies on a set of conditions that determine whether the ML model's prediction should be accepted, or bypassed in favor of a full exploratory optimization. There are three such bypass conditions: 
\begin{enumerate}
    \item a mandatory warm-up period of duration $t_{\text{warmup}}$ that begins at the time of the last reset $t_{\text{reset}}$; 
    \item a periodic trigger activated every $t_{\text{periodic}}$ time steps; and 
    \item a model uncertainty check based on a generalized uncertainty metric $\Psi(\cdot)$ (e.g., standard deviation, entropy, or ensemble variance). This condition triggers if any target component $j$ exceeds its specific threshold: $\Psi_j(\phi_k) \ge \theta_{\text{unc}, j}$. This formulation is designed to function as a confident baseline check, where the baseline itself is a tunable hyperparameter established either empirically from historical data or from domain knowledge of the system's most frequent operational strategy.
\end{enumerate}

\paragraph{Main Control Loop}
The main loop (Algorithm \ref{alg:main_loop_general}) operates on a receding-horizon basis, repeatedly solving an optimization problem to determine the optimal control plan for the upcoming horizon. The central feature of our method lies in how this plan is determined at each cycle. The loop first invokes the ML-driven sub-routine (detailed below) to efficiently determine the terminal target state $\mathbf{v}_{\mathrm{target}}$. Using this target as a terminal cost, it then solves the detailed, high-resolution optimization problem to find the final control actions. This loop also incorporates an adaptive reset mechanism to handle major changes in market conditions. When a significant shift in the price forecast triggers a reset, indicative of a seasonal or market regime change, the system deliberately discards the elite pool. While caching solutions is effective for recurring daily patterns, solutions optimized for a previous operational regime (e.g., winter prices) are likely suboptimal for the new one and could negatively bias the search. Therefore, the reset boosts exploration budgets to $E_r^{\text{init}}$ and initiates a new warm-up period, forcing the ML model to adapt to the new environment. The elite pool $\mathbf{\mathcal{E}}$ and the training datasets $\{\mathcal{D}_j\}$ are cleared as well to avoid concept drift and bias towards prior seasons.

\paragraph{ML-Driven Terminal Target Determination} 
This sub-routine (Algorithm \ref{alg:determine_target_general}) is where the primary computational savings are realized. Its goal is to provide the terminal target vector $\mathbf{v}_{\mathrm{target}}$ for the two-stage multi-resolution optimization problem (described in next section). A feature vector $\phi_k$ is constructed from the current forecasts and fed to the online-trained ML model $\{\mathcal{M}_j\}$ which yield predictions and a generalized uncertainty estimate via $\Psi(\cdot)$. A decision is then made between two paths:

Predict Path (Default): The algorithm bypasses the expensive exploratory optimization and simply sets $\mathbf{v}_{\mathrm{target}} \gets \mathbf{\hat{v}}$, where $\mathbf{\hat{v}}$ is the vector that contains the mean predictions of the ML models specific for each terminal state.

Compute Path (Exception): If any of the bypass conditions defined in the Algorithm Parameters are met, a full, exploratory optimization ($r=\hat{e}$) is executed. This is done to gather new data during warm-up and periodic triggers or to ensure safety when the model's uncertainty is high. If this path is taken, the resulting optimal target vector is used, and crucially, the new observation pairs $(\phi_k, v_{\mathrm{target}, j})$ are added to their respective datasets $\mathcal{D}_j$ to continuously retrain and improve the ML models.

\paragraph{Two-Stage Optimization Cascade}
Once $\mathbf{v}_{\mathrm{target}}$ is determined, this sub-routine (Algorithm \ref{alg:find_control_general}) executes a two-stage optimization incorporating the $\mathbf{v}_{\mathrm{target}}$ as the terminal cost. First, a low-resolution ($r=l$) optimization is performed. Second, the final, high-resolution ($r=h$) stage is seeded with an initial set blended from two sources: $N_{\mathrm{elites}}$ solutions sampled from the global elite pool  set $\mathbf{\mathcal{E}}$ and the $N_{\mathrm{seed}}$ top-performing solutions from the low-resolution stage. The low-resolution solutions are up-sampled to the finer timescale using a linear interpolation operator, denoted by $\mathcal{I}_{l\to h}$.

\begin{algorithm}[H]
\caption{ML-Accelerated Optimal Control (Main Loop)}
\label{alg:main_loop_general}
\begin{algorithmic}[1]
\Require
ML models $\{\mathcal{M}_j\}_{j=1}^{N_v}$, total time $T_{\mathrm{end}}$, resolution parameters $\{\Delta t_r, H_r, \mathcal{S}_r\}$, budgets $\{E_r^{\text{init}}, E_r^{\text{def}}\}$, uncertainty thresholds $\{\theta_{\text{unc}, j}\}_{j=1}^{N_v}$, $t_{\text{warmup}}, t_{\text{periodic}}$.
\State Let $K_{\text{end}} = T_{\text{end}} / \Delta t_{\hat{h}}$
\For{$k \gets 0, H_{\hat{h}}, 2H_{\hat{h}}, \dots, K_{\text{end}}$}
    \State Obtain current state $x_k$ at time $t_k = k \cdot \Delta t_{\hat{h}}$
    \State Acquire price forecast for reset check: $\mathbf{p}_k^{\text{reset}}$
    
    \If{$k=0$ \textbf{or} $\|\mathbf{p}_k^{\text{reset}} - \mathbf{p}_{\text{last}}\|_\infty > 10^{-8}$} \Comment{Detect change in price profile}
        \State \textit{//-- Reset logic: Clear elites, clear datasets, boost budgets, and start new warm-up --//}
        \State $\mathbf{\mathcal{E}} \gets \emptyset$;\State $\{\mathcal{D}_j\} \gets \emptyset$; \Comment{Clear all datasets}
        $E_r \gets E_r^{\text{init}}$ for all $r$ 
        \State $\mathbf{p}_{\text{last}} \gets \mathbf{p}_k^{\text{reset}}$
        \State $t_{\text{reset}} \gets t_k$
        \State 
        
    \EndIf

    \State Acquire forecasts and form ML feature vector $\phi_k$
    
    \State $(\mathbf{v}_{\mathrm{target}}, \{\mathcal{D}_j\}, \{\mathcal{M}_j\}) \gets \texttt{DetermineTerminalTarget}(x_k, \phi_k, t_k, t_{\text{reset}}, \{\mathcal{M}_j\}, \{\mathcal{D}_j\}, \{\theta_{\text{unc}, j}\}_{j=1}^{N_v})$ \Comment{Pass/Return sets}

    \State $(U^{(\hat{h})*}, \mathcal{P}^{(\hat{h})*}) \gets \texttt{FindOptimalControl}(x_k, \mathbf{v}_{\mathrm{target}}, \mathbf{\mathcal{E}})$

    \State \textit{//-- Apply control and update algorithm state for next cycle --//}
    \State $\mathbf{U}_{\text{hist}} \gets \text{Concatenate}(\mathbf{U}_{\text{hist}}, U^{(\hat{h})*})$
    \State $\mathbf{\mathcal{E}} \gets \mathbf{\mathcal{E}} \cup \mathcal{P}^{(\hat{h})*}$

    \If{$E_{\hat{h}} = E_{\hat{h}}^{\text{init}}$} \Comment{If this was a boosted cycle...}
        \State $E_r \gets E_r^{\text{def}}$ for all $r$ \Comment{...reset budgets to default values.}
    \EndIf
\EndFor
\end{algorithmic}
\end{algorithm}

\begin{algorithm}[H]
\caption{Sub-routine: Determine Terminal Targets}
\label{alg:determine_target_general}
\begin{algorithmic}[1]
\Function{DetermineTerminalTargets}{$x_k, \phi_k, t_k, t_{\text{reset}}, \{\mathcal{M}_j\}, \{\mathcal{D}_j\}, \{\theta_{\text{unc}, j}\}_{j=1}^{N_v}$} \Comment{Modified: Args}
    \State \textit{//-- Decide whether to run expensive optimization or trust the ML models --//}
    \State is\_warmup $\gets (t_k - t_{\text{reset}}) < t_{\text{warmup}}$
    \State is\_periodic\_trigger $\gets t_k \pmod{t_{\text{periodic}}} = 0$
    \State is\_uncertain $\gets$ \textbf{false}
    \State Initialize prediction vector $\mathbf{\hat{v}} \in \mathbb{R}^{N_v}$
    \For{$j \gets 1$ \textbf{to} $N_v$}
        \State $\mathbf{\hat{v}}[j] \gets \text{Predict}(\mathcal{M}_j, \phi_k)$
        \State $u_{\text{val}} \gets \Psi_j(\mathcal{M}_j, \phi_k)$ \Comment{Calculate generic uncertainty}
        \If{$u_{\text{val}} \ge \theta_{\text{unc}, j}$} 
            \State is\_uncertain $\gets$ \textbf{true}
        \EndIf
    \EndFor
    
    \If{is\_warmup \textbf{or} is\_periodic\_trigger \textbf{or} is\_uncertain}
        \State \textit{//-- Compute Path: Run full exploratory optimization --//}
        \State $[U^{(\hat{e})*}, X^{(\hat{e})*}] \gets \text{Solve}(\arg\min_{U^{(\hat{e})}} \sum_{i=0}^{H_{\hat{e}}-1} \ell(x_i, u_i, t_i) \text{ s.t. } x_0=x_k)$
        \State $T_{\text{target\_horizon}} \gets H_{\hat{l}} \cdot \Delta t_{\hat{l}}$
        \State $x_{\text{target}} \gets \text{State from trajectory } X^{(\hat{e})*} \text{ at time } t_k + T_{\text{target\_horizon}}$
        \State $\mathbf{v}_{\mathrm{target}} \gets \text{ExtractTargetComponents}(x_{\text{target}})$
        
        \State \textit{//-- Online learning: Update datasets and retrain the models --//}
        \For{$j \gets 1$ \textbf{to} $N_v$} \Comment{Loop to update specific datasets}
            \State $\mathcal{D}_j \gets \mathcal{D}_j \cup \{(\phi_k, \mathbf{v}_{\mathrm{target}}[j])\}$
        \EndFor
        \State $\{\mathcal{M}_j\} \gets \text{RetrainModels}(\{\mathcal{D}_j\})$
    \Else
        \State \textit{//-- Predict Path: Use the confident ML prediction --//}
        \State $\mathbf{v}_{\mathrm{target}} \gets \mathbf{\hat{v}}$
    \EndIf
    \State \Return $(\mathbf{v}_{\mathrm{target}}, \{\mathcal{D}_j\}, \{\mathcal{M}_j\})$ \Comment{Return sets}
\EndFunction
\end{algorithmic}
\end{algorithm}

\begin{algorithm}[H]
\caption{Sub-routine: Two-Stage Optimal Control Cascade}
\label{alg:find_control_general}
\begin{algorithmic}[1]
\Function{FindOptimalControl}{$x_k, \mathbf{v}_{\mathrm{target}}, \mathbf{\mathcal{E}}$}
    \State \textit{//-- Stage 1: Low-Resolution Optimization ($r=\hat{l}$) --//}
    \State $(U^{(\hat{l})*}, \mathcal{P}^{(\hat{l})*}) \gets \text{Solve for the solution to:}$
    \Statex \hspace{\algorithmicindent} $\arg\min_{U^{(\hat{l})}} \left[ \sum_{i=0}^{H_{\hat{l}}-1} \ell(x_i, u_i, t_i) + V_f^{(\hat{l})}(x_{H_{\hat{l}}}, \mathbf{v}_{\mathrm{target}}) \right]$

    \State \textit{//-- Stage 2, Step A: Blend initial set for high-resolution --//}
   \State $N_{\text{elites}} \gets \min(|\mathbf{\mathcal{E}}|, \lfloor\mathcal{S}_{\hat{h}}/2\rfloor)$
    \State $\mathbf{\mathcal{E}}_{\text{sub}} \gets \text{UniformRandomSample}(\mathbf{\mathcal{E}}, N_{\text{elites}})$
    
    \State $\mathcal{P}_{\text{interp}} \gets \{ \mathcal{I}_{\hat{l}\to\hat{h}}(U) \mid U \in \mathcal{P}^{(\hat{l})*} \}$ \Comment{$\mathcal{I}$ is linear interpolation}
    \State Sort $\mathcal{P}_{\text{interp}}$ by cost
    \State $N_{\text{seed}} \gets \mathcal{S}_{\hat{h}} - N_{\text{elites}}$
    \State $\mathcal{P}_{\text{seed, sub}} \gets \text{first } N_{\text{seed}} \text{ solutions from } \mathcal{P}_{\text{interp}}$
    
    \State $\mathcal{P}_0^{(\hat{h})} \gets \mathbf{\mathcal{E}}_{\text{sub}} \cup \mathcal{P}_{\text{seed, sub}}$ \Comment{Final blended initial set}

    \State \textit{//-- Stage 2, Step B: High-Resolution Optimization ($r=\hat{h}$) --//}
        \State $(U^{(\hat{h})*}, \mathcal{P}^{(\hat{h})*}) \gets \text{Solve for the solution to:}$
        \Statex \hspace{\algorithmicindent} $\arg\min_{U^{(\hat{h})}} \left[ \sum_{i=0}^{H_{\hat{h}}-1} \ell(x_i, u_i, t_i) + V_f^{(\hat{h})}(x_{H_{\hat{h}}}, \mathbf{v}_{\mathrm{target}}) \right]$
        \Statex \hspace{\algorithmicindent} using initial set $\mathcal{P}_0^{(\hat{h})}$

    \State \Return $(U^{(\hat{h})*}, \mathcal{P}^{(\hat{h})*})$
\EndFunction
\end{algorithmic}
\end{algorithm}

\subsubsection{Pilot System Implementation}

This section provides the specific details of applying the general framework to the Pilot System involving a heat pump and a battery. In the context of our system, the generic target vector $\mathbf{v}_{\mathrm{target}}$ from the algorithm contains the battery's State of Charge (SoC), $\mathrm{SoC}_{\mathrm{target}}$ and the median temperature of the thermal storage $\mathrm{T_{median,target}}$. These values guide the trade-off between completely emptying the battery and the thermal storage at the end of the time period for immediate cost savings within the $H_{\hat{l}}$ and $H_{\hat{h}}$ horizons versus preserving some of its charge for potentially more valuable uses in the future, as determined by the long-horizon exploratory optimization. The $\text{ExtractTargetComponent}(x)$ function is therefore implemented as a simple selection of the SoC and median thermal storage temperature components from the state vector $x$. 

\paragraph{System Model and Control Spaces}

The state vector $x_k \in \mathbb{R}^{n_r}$ includes the battery's SoC $[x_k]_{SoC}$, the thermal storage median temperature $[x_k]_{T_{\mathrm{median}}}$, and the instantaneous energy production output $[x_k]_{\mathrm{q_{prod}}}$. The control input vector for this system is defined as $u_k = [u^{\mathrm{pum}}_k, u^{\mathrm{boi}}_k, u^{\mathrm{bat}}_k]^\top$. The pump commands for the heat pump and boiler, $[u^{\mathrm{pum}}_k, u^{\mathrm{boi}}_k]^\top \in [0, u^{\mathrm{nom}}]^2$, represent their respective mass flow rates measured upstream of the valve, and $u^{\mathrm{nom}}$ is the nominal mass flow rate. The battery command, $u^{\mathrm{bat}}_k \in [-1, 1]$, is a normalized, dimensionless variable representing charge/discharge instructions.

\paragraph{Control Projection Rules}
The unconstrained decision variables are mapped to physical commands through the following projection rules. For both pumps, a minimum flow rate is enforced to prevent inefficient operation at low part load ratio as
\begin{equation}
u^{\mathrm{pum,proj}}_k =
\begin{cases}
0, & \text{if } u^{\mathrm{pum}}_k < \alpha_{nom} \, u^{\mathrm{nom}} \\
u^{\mathrm{pum}}_k, & \text{otherwise,}
\end{cases}
\end{equation}
where $\alpha_{nom}$ is the minimum part load operation.

The normalized battery command is projected to a physical power command $P^{\mathrm{bat,proj}}_k$. This projection accounts for the available energy based on the current SoC, battery capacity $E_{\max}$, discharging and charging efficiencies $\eta_{discharge}$ and $\eta_{charge}$, and the battery's nominal power rating $P_{\mathrm{bat}}^{\mathrm{nom}}$ as
\begin{equation}
P^{\mathrm{bat,avail}}_k =
\begin{cases}
\dfrac{(1 - \mathrm{SoC}_k) \, E_{\max}}{\Delta \, t_r\eta_{charge}}, & \text{if } u^{\mathrm{bat}}_k \ge 0 \quad (\text{charging}) \\[2em]
\dfrac{\mathrm{SoC}_k \, E_{\max} \, \eta_{discharge}}{\Delta t_r}, & \text{if } u^{\mathrm{bat}}_k < 0 \quad (\text{discharging}).
\end{cases}
\end{equation}

This available power is then scaled by the normalized command and clipped by the battery's nominal power rating $P_{\mathrm{bat}}^{\mathrm{nom}}$ which defines the absolute upper and lower power limits as
\begin{equation}
P^{\mathrm{bat,proj}}_k = \mathrm{clip}\left(u^{\mathrm{bat}}_k \, P^{\mathrm{bat,avail}}_k, -P_{\mathrm{bat}}^{\mathrm{nom}}, P_{\mathrm{bat}}^{\mathrm{nom}}\right).
\end{equation}
This ensures the commanded power respects both state-dependent energy limits and the absolute power limits.

\paragraph{Cost Function}
This section defines the specific components that constitute the stage cost $\ell$ and the terminal cost $V_f$ in our problem formulation.

\textbf{Stage Cost $\ell$} \\
The stage cost $\ell(x_k, u_k, w_k)$ represents the total cost incurred during each time step $k$. It is the sum of the direct energy cost and a penalty for daytime production shortfall:

\begin{enumerate}
    \item The \textbf{energy cost} $\ell_{\mathrm{energy}}$ is the incremental energy cost computed by the plant model based on known price $p(t_k)$ and weather $w(t_k)$ forecasts:
    \begin{equation}
    \ell_{\mathrm{energy}}(x_k, u_k, t_k) = C_{\mathrm{energy}}\big(x_k, u_k, p(t_k), w(t_k)\big)
    \end{equation}

    \item The \textbf{production shortfall penalty} $\ell_{\mathrm{shortfall}}$ penalizes any failure to meet the thermal production target $q_{\mathrm{prod,target}}$ required to meet the plant load demand during the specific hours the plant is in operation. Rather than representing a physical operating cost, this term serves as a feasibility-enforcing penalty that penalizes any candidate trajectory during optimization in which the thermal production falls short of the required target, effectively ensuring that only feasible solutions are retained in the final result. It is
    \begin{equation}
    \ell_{\mathrm{shortfall}}(t_k, [x_k]_{\mathrm{q_{prod}}}) = 
    \begin{cases}
    \gamma_{\mathrm{day}} \,\max\!\big(0,\, q_{\mathrm{prod,target}} - [x_k]_{\mathrm{q_{prod}}}\big), & \text{if } t_k \in [08{:}00,\,20{:}00],\\
    0, & \text{otherwise,}
    \end{cases}
    \end{equation}
\end{enumerate}
where $\gamma_{\mathrm{day}} > 0$ is a penalty term.
The total stage cost included in the main objective function's summation is thus $\ell(x_k, u_k, w_k) = \ell_{\mathrm{energy}}(x_k, u_k, w_k) + \ell_{\mathrm{shortfall}}(x_k, u_k, w_k)$.

\vspace{\baselineskip}
\vspace{\baselineskip}
\textbf{Terminal Cost $V_f$} \\
The terminal cost $V_f(x_H)$ is a soft constraint applied only once to the final state $x_H$. Its purpose is to guide the system towards desirable terminal states. It measures the deviation of the final SoC and thermal storage temperature from their respective target values, $\mathrm{SoC}_{\mathrm{target}}$ and $T_{\mathrm{median, target}}$. These targets are the key outputs from the initial exploratory optimization and provide consistent long-term goals for subsequent refinement stages. The terminal cost is formulated as the weighted sum of these deviations:
\begin{equation}
V_f(x_H) =
\begin{cases}
\gamma_{\mathrm{soc}}\,\big|[x_{H_r}]_{\mathrm{SoC}} - \mathrm{SoC}_{\mathrm{target}}\big| + \gamma_{T}\,\big|[x_{H_r}]_{T_{\mathrm{median}}} - T_{\mathrm{median, target}}\big|, & \text{if } r \in \{\hat{l}, \hat{h}\}, \\
0, & \text{if } r = \hat{e}.
\end{cases}
\end{equation}

With the production shortfall measured in megawatts (MW), the $\gamma_{\mathrm{day}}$ factor applies a cost of 1000 for every 1 MW of unmet daytime demand. Note that the exploratory stage itself ($r = \hat{e}$) does not use the terminal cost term ($V_f = 0$), as its objective is to determine the optimal $\mathrm{SoC}_{\mathrm{target}}$ and $T_{\mathrm{median, target}}$. To address the disparity in magnitude between the tank temperature $T_{\text{median}}$ and the $SoC$, the corresponding penalty weights were calibrated via numerical experimentation to ensure a balanced contribution to the cost function.

\paragraph{Multi-Resolution Scheme}
The optimization proceeds sequentially through three stages:
\begin{itemize}
\item \textbf{Exploratory Stage ($r = \hat{e}$):} A 48-hour horizon with a 2-hour time step ($H_{\hat{e}} = 24$). This stage runs without the terminal SoC and $T_{\text{median}}$ penalties ($\gamma_{\mathrm{soc}} = 0$ and $\gamma_{\mathrm{T}} = 0$). This extended horizon is critical for avoiding myopic decisions, as it allows the optimizer to properly value the ``carryover" of stored energy from day 1 into day 2. Its purpose is to determine the optimal $\mathrm{SoC}_{\mathrm{target}}$ and $T_{\mathrm{median, target}}$ values at the 24-hour mark, which is then passed to the subsequent stages. This is the stage that the ML model aims to replace.
\item \textbf{Low-Resolution Stage ($r = \hat{l}$):} A 24-hour horizon with a 1-hour time step ($H_{\hat{l}} = 24$). It uses the $\mathrm{SoC}_{\mathrm{target}}$ and $T_{\mathrm{median, target}}$ values from the exploratory stage to compute its terminal penalty. This 24-hour duration represents a crucial trade-off: it is long enough to encompass a full diurnal cycle of solar PV generation and plant load, it provides complete daily solutions suitable for populating the elite pool for recurring daily patterns, and it keeps the problem computationally tractable.
\item \textbf{High-Resolution Stage ($r = \hat{h}$):} A 24-hour horizon with a 30-minute time step ($H_{\hat{h}} = 48$). It also uses the $\mathrm{SoC}_{\mathrm{target}}$ and $T_{\mathrm{median, target}}$  values, and its solution provides the actual controls to be applied to the system. The 24-hour horizon is maintained for the reasons stated above, while the fine-grained time step is necessary to better align the rate of change of control action with the major system dynamics.
\end{itemize}

\paragraph{Algorithm Hyperparameters}
\begin{itemize}
    \item \textbf{Black-Box Optimizer:} The proposed framework is solver-agnostic. We use the Adaptive Differential Evolution algorithm (JADE) \cite{zhang2009jade}. Key hyperparameters include a mutation greediness $p=0.05$, parameter adaptation rate $c=0.1$, and the ``DE/current-to-pbest" strategy. Population sizes were set to $\mathcal{S}_{\hat{e}}=48$, $\mathcal{S}_{\hat{l}}=96$, and $\mathcal{S}_{\hat{h}}=96$. We utilize our own implementation of JADE in \texttt{PyTorch 2.3.0} and \texttt{Python 3.11} \cite{paszke2019pytorch}. The JADE solver is openly available as a part of the whole framework.
    \item \textbf{Budgets:} Default optimization budgets were 4000 evaluations for the exploratory stage, and 5000 for low-resolution and high-resolution stages. During a reset, the high-resolution budget was boosted to $20,000$. 
    \item \textbf{Triggers:} For the SoC terminal target, the low uncertainty threshold of $\theta_{\text{unc}}=0.01$ was set, as it establishes a highly conservative trigger. This combination ensures that the computationally inexpensive ML path is only used when the ML model's upper confidence bound of the target SoC is itself below 1\%, effectively green-lighting only the most confident predictions of the common `discharge' strategy. For the thermal storage median temperature, the uncertainty threshold $\theta_{\text{unc}}$ was set to be $71^\circ \mathrm C$. This threshold was selected to force the system to run the full exploratory stage whenever the ML model gives any indication that there may be some benefit to the tank being partially charged at the end of the operating horizon; exploration is only avoided when the ML model is completely confident that the tank should be emptied (i.e., the temperature is at 70\textsuperscript{o}C, the allowable lower bound of operation). In essence, the thresholds act as a decision point: the algorithm assesses if it is confident enough to commit to the greedy discharge strategy based on current forecasts, or if the level of uncertainty warrants invoking the full exploratory optimizer to find a potentially more better solution.
    
    Finally, we set a warmup period of $t_{\text{warmup}}=240$ hours (10 days), and a periodic trigger of $t_{\text{periodic}}=120$ hours (5 days). 
    \item \textbf{Feature Vectors:} The price reset vector $\mathbf{p}_k^{\text{reset}}$ uses a 24-hour forecast at 30-minute intervals (48 dimensions). The ML input feature vector $\phi_k$ uses a 48-hour weather forecasts at 2-hour intervals (24 dimensions total), while the output features are the $\mathrm{SoC}_{\mathrm{target}}$ and the $\mathrm{T_{median}}$.
\end{itemize}

\subsubsection{ML Models}

We evaluated two distinct supervised learning ML architectures to generate the terminal target predictions $\mathbf{\hat{v}}$ and the associated uncertainty estimates $\Psi(\cdot)$ required by our framework. Both implementations utilize \texttt{scikit-learn 1.6.1} \cite{pedregosa2011scikit} and were executed with fixed random seeds to ensure a controlled experimental environment.

\textbf{Random Forest (RF):}
The first approach employs RF regressors \cite{breiman2001random}. We train one independent model with 200 decision trees for each target component $j$. This architecture is efficient as it allows for simultaneous estimation of the value and the uncertainty from a single model. For a given input $\phi_k$, the mean prediction ${\hat{v}}_j$ is computed as the average of the predictions from all individual trees in the specific model for target $j$. The uncertainty $\sigma_{k,j}$ is derived from the standard deviation of these individual tree predictions. The upper confidence bound used to trigger the uncertainty bypass condition is calculated assuming a Gaussian distribution of errors as $\text{Upper Bound} = \mathbf{\hat{v}}_j + \kappa \cdot \sigma_{k,j}$. We evaluated this approach using two distinct exploration settings: $\kappa=1$ and $\kappa=2$, corresponding to theoretical confidence intervals of approximately 68\% and 95\%, respectively, under the normality assumption.

\textbf{Gradient Boosting with Quantile Regression (GB):}
The second approach utilizes GB Regressors. Unlike the RF approach, this method requires training two distinct models for each target component $j$. The first model is trained to predict the mean value ${\hat{v}}_j$ using the standard squared error loss. The second model is specifically trained to estimate the upper confidence bound directly using quantile regression. This second model minimizes the pinball loss function $L_\alpha$. Defining $z$ as the true target value and $\hat{z}$ as the predicted value, the loss for a target quantile $\alpha=0.95$ is:
\begin{equation}
L_\alpha(z, \hat{z}) = \max(\alpha (z - \hat{z}), (\alpha - 1) (z - \hat{z}))
\end{equation}
Because $\alpha$ is close to 1, this loss function imposes a heavy penalty on underestimation ($z > \hat{z}$) and a small penalty on overestimation. Consequently, the model learns to output a prediction $\hat{z}$ that exceeds the true value $z$ in approximately 95\% of cases.
Physically, this intentional overestimation of terminal targets creates a conservative buffer. While this forces the optimizer to perform a more exhaustive and computationally expensive search to meet stricter targets, it safeguards against the sub-optimality that would arise from underestimating the true terminal requirements.

While the RF approach is computationally lighter (one model per target), the calculation of the upper bound using $\kappa \cdot \sigma$ implicitly assumes that the posterior distribution of the target uncertainty is Gaussian. The GB approach with quantile loss is theoretically superior for this application because it is distribution-free; it does not rely on assumptions of normality. This allows the framework to more accurately capture the asymmetric or heavy-tailed risk profiles often seen in energy systems, albeit at the cost of managing and training twice as many models.


\section{Results}\label{sec: results}
In this section, we present the results for the pilot case study. We start with a brief discussion of the results from the architectural optimization process (Section \ref{subsec:arch-opt-res}). We then provide a detailed analysis of the limit of performance analysis, presenting results that highlight the benefits of both the multi-fidelity and ML contributions (Sections \ref{subsec:lop-analysis}-\ref{subsec:ablation}). In Section \ref{subsec:op-behaviour-comps}, we present a brief comparison of the actual control decisions produced by the different strategies explored in this work.

For our analysis, we consider electricity prices in California obtained through the California Energy Commission's Market Informed Demand Automation Server~\cite{calflexhub}. Details on the electricity prices, along with the weather data and equipment design parameters used in our case study, are presented in Appendix \ref{app:simulation-data}.  Equipment cost data for turn-key installed components are presented in Table \ref{tab:cap-cost-tables}.

\subsection{Architectural Optimization Results}\label{subsec:arch-opt-res}
\begin{figure}
    \centering
    \includegraphics[width=0.75\linewidth]{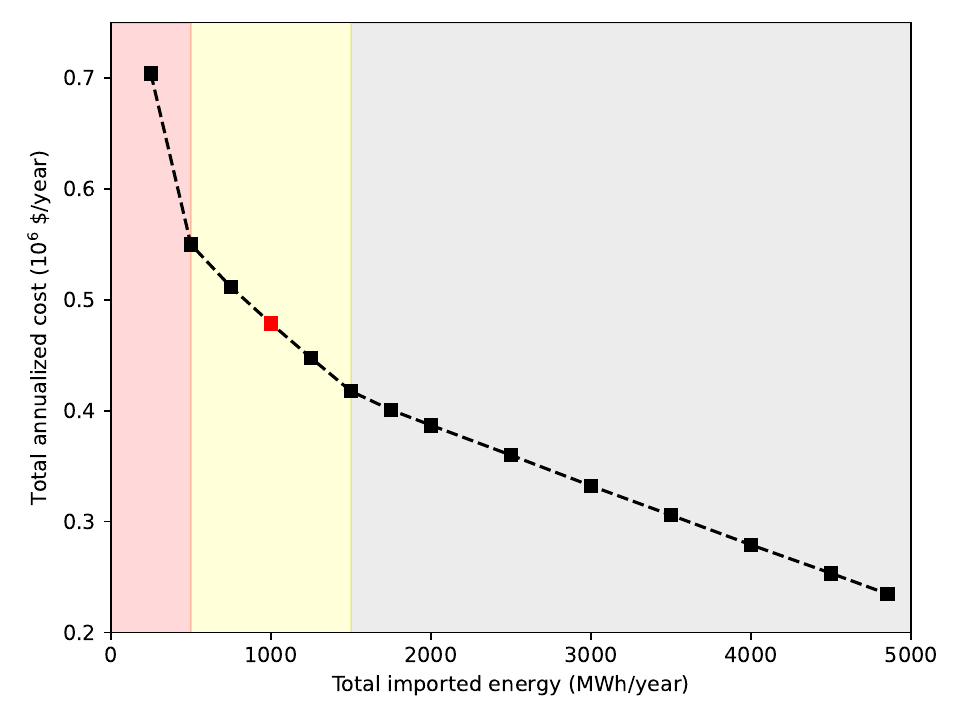}
    \caption{Pareto front for multi-objective problem. The squares represent individual designs on the solution front. The shaded areas highlight the three distinct regimes observed. The red square represents the design to be considered in the limit of performance analysis.}

    \label{fig:pareto-front}
\end{figure}

Figure \ref{fig:pareto-front} shows the Pareto front obtained for the architectural design optimization problem. Limiting the amount of energy that can be imported from outside the facility incurs a significant cost penalty; the TAC of the designs range from to $\mathrm{\unitfrac[\$0.235-0.704]{M}{yr}}$; a three-fold cost increase. 

\begin{figure}
    \centering
    \includegraphics[width=0.75\linewidth]{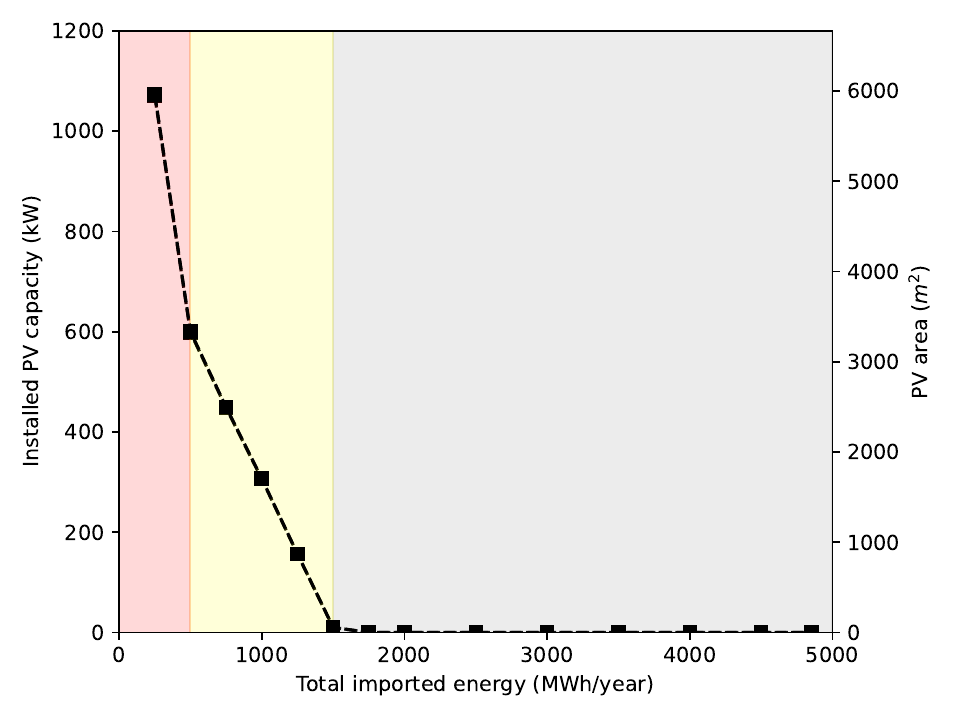}
    \caption{Installed PV capacities for designs on Pareto front.}
    \label{fig:pv-area-plot}
\end{figure}

The Pareto front exhibits three distinct regimes (shaded areas), with the gradient of the Pareto front decreasing in each regime (left to right). To better understand the shape of the Pareto front, we present a plot of the installed PV capacities in Figure \ref{fig:pv-area-plot}, and plots of the generation and storage capacities of the designs are presented in Appendix \ref{app:pareto-analysis}. The shape of the trade-off curve is clearly driven by photovoltaic installation.
The rightmost regime (gray) corresponds to the regime with no PV installation; all energy demands from the plant are met by either purchasing electricity from the grid or purchasing gas. In this regime ($\mathrm{TE}>\unitfrac[1500]{MWh}{yr}$), the Pareto front is linear and shows the slowest rate of cost increase ($\approx\$53$ per MWh of imported energy reduction). In the far-right design ($\mathrm{TE}=\unitfrac[4852]{MWh}{yr};\mathrm{TAC=\unitfrac[\$0.235]{M}{yr}}$)  which represents the design when there is essentially no constraint on energy imports, the $\mathrm{TE}$ is just above the annual thermal demand of the plant ($\unitfrac[4380]{MWh}{yr}$). The design uses only natural gas for heat generation; a 1MW gas boiler is installed and operated continuously to meet the load demands (no storage required). Designs in the gray regime trade off between the installed heat pump and gas boiler capacities while continuously increasing the installed tank and battery storage capacities; the leftmost design in the gray regime drops the gas boiler to near zero. Beyond this point, PVs are installed to reduce the imported energy. The maximums of the installed storage capacities for both the battery ($\unit[1.2]{MWh}$) and hot water tank ($\unit[2.0]{MWh}$) occur in this regime.

For designs in the yellow regime $\left(\unitfrac[500]{MWh}{yr}<\mathrm{TE}<\unitfrac[1500]{MWh}{yr}\right)$, PV installation is required. Designs in the yellow shaded region are solely dependent on heat pumps for heat generation alongside battery and hot water storage; gas boiler capacities are less than 1\% of the plant load ($<\unit[10]{kW}$). Designs in this regime come at a cost penalty, costing roughly twice as much as in the gray regime ($\approx\$120-130$ per MWh of imported energy reduction). 

For designs in the red regime ($\mathrm{TE}<\unitfrac[500]{MWh}{yr}$), every additional MWh of imported energy reduction comes at a steep cost. For the left-most design ($\mathrm{TE}=\unitfrac[250]{MWh}{yr}$), the optimal solution is to install PV and heat pump capacities of $\unit[1.07]{MW}$ and $\unit[1.00]{MW}$, respectively, roughly the level of the hourly load. 

\begin{table}
\caption{Equipment sizes for design selected for the limit of performance analysis{*} \label{tab:Equipment-sizes-forlop-analysis}}

\begin{centering}
\begin{tabular}{cr}
\toprule 
Component & Capacity\tabularnewline
\midrule
PV & 306.2 kW\tabularnewline
Heat pump & 862.2 kW\tabularnewline
Battery & 1124.8 kWh\tabularnewline
Hot water tank & 1911 kWh\tabularnewline
\bottomrule
\end{tabular}
\par\end{centering}
\centering{}{*}The optimal design includes a 6kW gas boiler, but since this is less then 1\% of the heat pump capacity, therefore we will in the limit of performance analysis and in the control verification below assume that the boiler contribution is negligible, i.e., $[u^{\mathrm{boi}}_k]^\top = 0$ for all $k$.
\end{table}

The results of the architectural optimization provides us with information on the optimal system configurations and equipment sizes required to meet the load of the pilot plant. Given the system architecture, the next step is design verification, where the goal is to design the best possible control policy/strategy for operating a fixed energy system architecture at a minimum cost while still satisfying all requirements. In selecting a design for the limit-of-performance analysis, we selected a design that includes some PV. The intrinsic variability of PV generation provides a valuable source of disturbance, enabling a more rigorous evaluation and stress-testing of our controller design methodologies. Based on this requirement, we select the design at the midpoint of the yellow regime ($\mathrm{TE}=\unitfrac[1000]{MWh}{yr};\mathrm{TAC=\unitfrac[\$0.479]{M}{yr}}$). Table \ref{tab:Equipment-sizes-forlop-analysis} summarizes the equipment sizes for the design. 

For our application, the modelica models represent
the heating and electrical system shown in Figure~\ref{fig:plant}. 
We used the optimization results from IDAES for the sizes of the heat pump, the PV, the battery and the hot water tank. As IDAES sized the gas boiler to less than 1\% of the size 
of the heat pump, we added a control option to permanently disable the gas boiler,
as this equipment will not be part of the recommended design.

\subsection{Limit of Performance Analysis}\label{subsec:lop-analysis}

In this section, we present the limit of performance analysis that uses our ML-Accelerated optimal control approach, which will be compared against a Rule-based controller. This comparison will show the performance gap between these two controllers. The primary metric for evaluation is the total cost of operating the system at the end of one year. To provide a comprehensive assessment, we deploy three distinct variants of the ML-Accelerated framework, each distinguished by its underlying predictive model and uncertainty quantification mechanism. We denote the RF-based implementations as $\mathrm{ML_{RF1}\text{-}MR_{ws}}$ and $\mathrm{ML_{RF2}\text{-}MR_{ws}}$, utilizing uncertainty weights of $\kappa=1$ and $\kappa=2$, respectively. Additionally, we evaluate $\mathrm{ML_{GB}\text{-}MR_{ws}}$, which employs the GB model calibrated to the 0.95 quantile loss.

Additionally, we assess the ML-Accelerated optimal control approaches against a multi-resolution strategy (denoted $\text{MR}_{ws}$) that incorporates elite pool warm starting but does not include ML acceleration. This will allow us to evaluate the computational savings and quantify any potential loss of solution quality due the ML model integration. We also compare it to a baseline single high-resolution approach (denoted as $\text{HR}$) that lacks elite pool warm starting. In the $\text{HR}$ approach, we use the same number of evaluations as all the other approaches, however, all are utilized to evaluate only the high-resolution objective function. Both the multi-resolution and high-resolution methods encompass the full exploratory stage, where we solve the exploratory optimization problem to obtain the terminal SoC. This analysis highlights how the ML-Accelerated approaches can achieve the same or even better performance while requiring a reduced exploratory evaluation budget over the entire year. For the three methods, all numerical experiments were repeated 4 times and conducted on a single AMD EPYC 9534 processor with 128 cores, utilizing parallel objective function evaluations. The comprehensive yearly optimal control run, considering the full evaluation budget of both high-resolution and multi-resolution approaches, required approximately 2.5 days to complete.

Drawing from the insights of the architectural optimization, we define the specific parameters for the limit of performance analysis. The boiler was determined to be insignificant and is therefore excluded from the problem, simplifying the control input vector to two dimensions, i.e., the heat pump mass flow rate and battery command. Consequently, the decision vectors for the exploratory, low-resolution, and high-resolution stages are defined as $U^{(\hat{e})} \in \mathbb{R}^{48}$, $U^{(\hat{l})} \in \mathbb{R}^{48}$, and $U^{(\hat{h})} \in \mathbb{R}^{96}$, respectively.

The key system parameters are set as follows: the battery has a maximum capacity of $E_{max} = 1124.8$ kWh and a nominal power rating of $P_{\mathrm{bat}}^{\mathrm{nom}} = 400$ kW, while the heat pump has a maximum mass flow rate of $u^{\mathrm{nom}} = 12$ kg/s (computed from water requirement to meet 1MW thermal load). For the control projection rules, a threshold of $\alpha_{nom} = 0.2$ is applied to the pump command. The weighting factors for the terminal cost penalty were determined through numerical tuning to be $\gamma_{\mathrm{day}} = 1000$, $\gamma_{\mathrm{soc}} = 5000$ and $\gamma_{T} = 5$. All optimizations utilize the same price $p(t_k)$ and weather $w(t_k)$ forecast data as the architectural optimization study, with further details provided in Appendix \ref{app:simulation-data}.

Figure \ref{fig:cost} illustrates the mean total operating cost of the system over the course of a full year for all evaluated methods (note that the IDAES and Rule-based approaches are only a single run). Firstly, the results indicate that the $\text{HR}$ method, specifically the out-of-the-box JADE solver, exhibits the poorest performance, falling significantly short of the Rule-based approach, with a mean total cost of \$96,225 at the end of the year. This underperformance of the $\text{HR}$ method underscores the need for developing more sophisticated and complex methodologies. Furthermore, the standard multi-resolution method $\mathrm{MR_{ws}}$ and the three ML-Accelerated variants demonstrate comparable performance, achieving mean total costs ranging between \$73,054 and \$74,157. All multi-resolution optimization-based approaches consistently outperform the Rule-based baseline, which incurred a total cost of \$80,194. Notably, while $\mathrm{ML_{RF2}\text{-}MR_{ws}}$ achieved the lowest average cost, the absolute best individual result across all experimental runs was achieved by the $\mathrm{ML_{GB}\text{-}MR_{ws}}$ method, with a minimum total cost of \$71,778. This yields a savings of \$8,416, representing a 10.5\% improvement over the Rule-based approach. {The best performing ML run produces significant controller performance improvement,  cutting the performance gap between the Rule-based controller and the theoretical lower bound established by the IDAES approach (\$60,284) by 42\%. 

The convergence of all stochastic methods remains relatively stable over 4 runs. As shown in Table \ref{tab:cost_statistics}, the standard deviations of the total annual operating cost from all methods is under 2\%, with  $\mathrm{MR_{ws}}$ and $\mathrm{HR}$ showing the highest and lowest percentage deviations, respectively. The GB approach $\mathrm{ML_{GB}\text{-}MR_{ws}}$, which produces the best single run of the ML-based approaches, also shows slightly higher variation than the other two AI-enabled approaches, though this remains comparable to the baseline multi-resolution strategy.

\begin{figure}[H]
    \centering
    \includegraphics[width=0.75\linewidth]{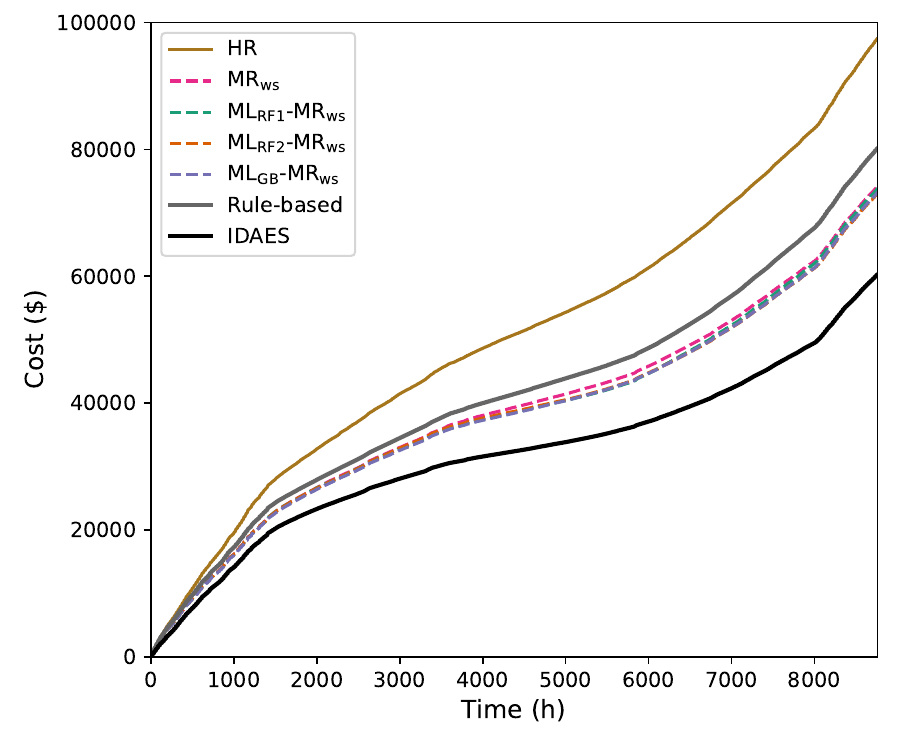}
    \caption{Mean total operating cost of the energy system over one year for all used methods. A lower operating cost at any point in the year indicates better performance.}
    \label{fig:cost}
\end{figure}

\begin{table}[htbp]
\centering
\caption{Total yearly cost statistics for different approaches over 4 runs. The standard deviation percentages are computed relative to the mean values.}
\label{tab:cost_statistics}
\begin{tabular}{lrrrrrr}
\hline
Approach & Mean (\$) & Median (\$) & Min (\$) & Max (\$) & Std (\$) & Std (\%) \\
\hline
$\mathrm{HR}$ & 97,482 & 97,498 & 97,019 & 97,913 & 317 & 0.33\\
$\mathrm{MR_{ws}}$ & 74,157 & 73,845 & 72,723 & 76,216 & 1,275 & 1.72 \\
$\mathrm{ML_{RF1}\text{-}MR_{ws}}$ & 73,807 & 74,024 & 72,660 & 74,520 & 695 & 0.95 \\
$\mathrm{ML_{RF2}\text{-}MR_{ws}}$ & 73,054 & 73,250 & 72,039 & 73,677 & 659 & 0.91\\
$\mathrm{ML_{GB}\text{-}MR_{ws}}$ & 73,188 & 73,109 & 71,778 & 74,758 & 1,075 & 1.47\\
Rule-based$^*$ & 80,194 & -- & -- & -- & -- & --\\
IDAES$^*$ & 60,284 & -- & -- & -- & -- & -- \\
\hline
\multicolumn{6}{l}{$^*$Single run (no statistics available)} \\
\end{tabular}
\end{table}


The results above indicate that all multi-resolution approaches outperform direct high-resolution optimization, regardless of whether the ML component is incorporated; importantly, the inclusion of the ML models do not degrade solution quality and appear to enhance performance stability across repeated runs. However, where the ML-Accelerated approaches really excel over the baseline multi-resolution method $\mathrm{MR_{ws}}$ is their ability to reduce the total number of evaluations during the exploratory search. Figure \ref{fig:evals_hist} illustrates the number of exploratory evaluations for each run across all variants.
All three ML-driven variants demonstrate a significant advantage over the $\mathrm{MR_{ws}}$ approach, which requires a fixed total of 1,460,000 exploratory evaluations. However, a comparison of the variants reveals that the GB approach $\mathrm{ML_{GB}\text{-}MR_{ws}}$ is the most computationally efficient. As shown in Table \ref{tab:evals_combined}, the GB method achieves a mean total of 962,000 evaluations, representing a 34\% reduction in function evaluation calls without any noticeable loss in solution quality. This is also  notably lower than the RF variants, which both average over 1.1 million evaluations. This efficiency stems from the GB model's specific uncertainty quantification; it triggers the expensive high upper confidence bound fallback condition significantly less often (averaging 498,000 triggers) compared to the RF models (averaging over 640,000 triggers). This suggests that the distribution-free quantile regression provides a more precise uncertainty bound than the Gaussian assumption used in the RF approaches, allowing the system to trust the ML predictions more frequently. Moreover, Comparing the two Random Forest variants reveals that the uncertainty scaling parameter $\kappa$ has a negligible impact on computational effort. The mean difference between $\mathrm{ML_{RF1}\text{-}MR_{ws}}$ ($\kappa=1$) and $\mathrm{ML_{RF2}\text{-}MR_{ws}}$ ($\kappa=2$) is approximately 1\%, indicating that the uncertainty estimates rarely fall in the marginal zone between one and two standard deviations. Thus, tuning $\kappa$ yields minimal efficiency gains compared to switching to the GB architecture, which significantly alters the evaluation frequency. Despite this, both RF-based approaches still manage to reduce the required number of evaluations by over 24\%, representing significant computational savings.

The standard deviation observed within each method is driven entirely by the stochastic nature of the exploration triggers, as the remaining evaluations are fixed overhead. This variability arises from the JADE solver used in the exploratory optimization, which generates slightly different optimal trajectories for the training targets in each run. These variations lead to evolving datasets and differing uncertainty estimates $\sigma_k$ across runs.
Crucially, Figure \ref{fig:cost_hist} compares the yearly operating costs of these ML runs against the $\mathrm{MR_{ws}}$ baseline. This comparison confirms that the massive computational reductions achieved by the $\mathrm{ML_{GB}\text{-}MR_{ws}}$ variant do not compromise system performance. Despite performing significantly fewer evaluations than the baseline, the ML-accelerated strategy maintains comparable operating costs, validating it as the preferred choice for solving the limit of performance problem.

\begin{figure}[H]
	\centering
	\begin{subfigure}[b]{0.85\textwidth}
		\centering
		\includegraphics[width=\linewidth]{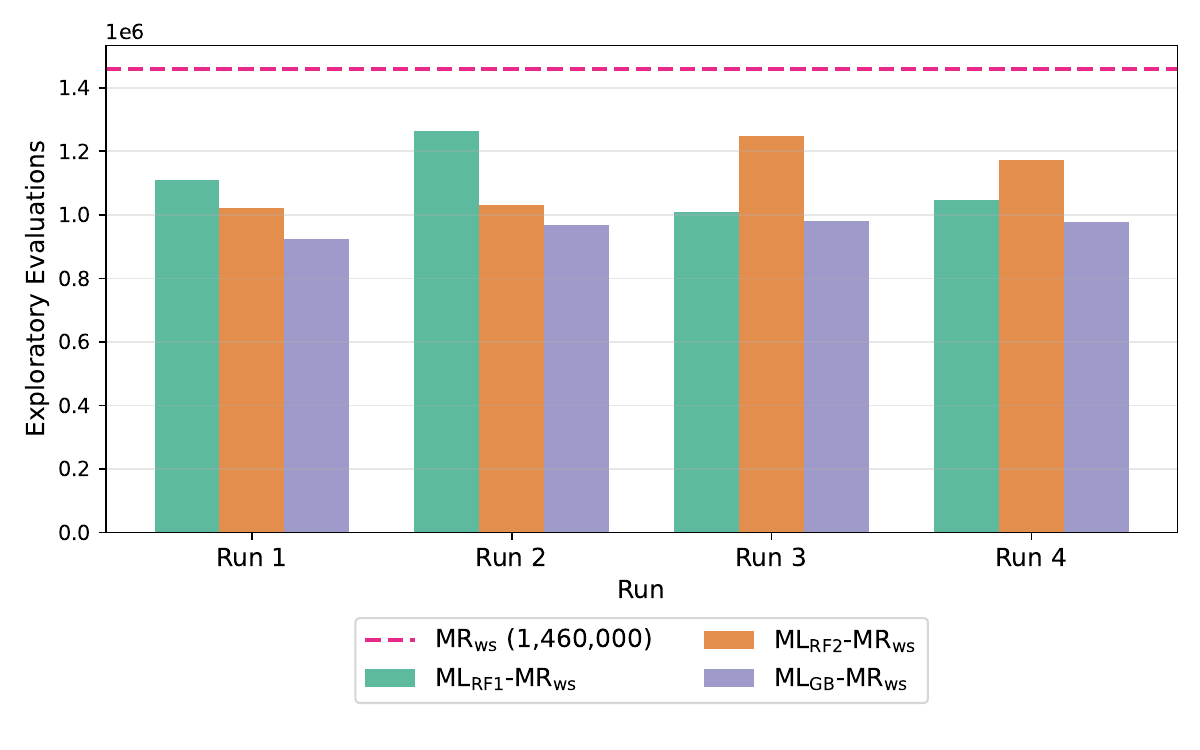}
		\caption{}
		\label{fig:evals_hist}
	\end{subfigure}
	\hfill 
	\begin{subfigure}[b]{0.85\textwidth}
		\centering
		\includegraphics[width=\linewidth]{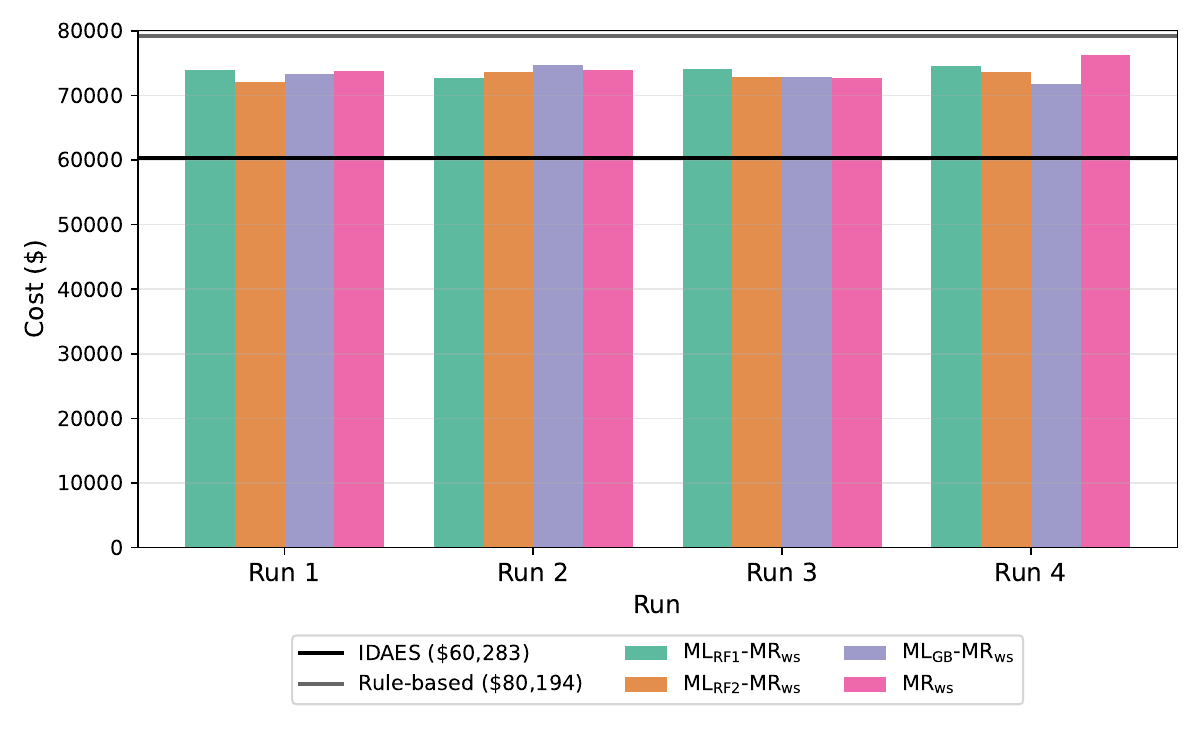}
		\caption{}
		\label{fig:cost_hist}
	\end{subfigure}
	\caption{A detailed analysis of each of the four runs of the ML-Accelerated variants is presented, where (a) displays the total number of exploratory evaluations (lower is better), and (b) shows a comparison of the total yearly operating costs (\$) of each run against the $\text{MR}_{ws}$ method and the best-performing IDAES approach (lower is better).}
	\label{fig:cost_evaluations_plot}
\end{figure}

\begin{table}[H]
\centering
\caption{Statistics of exploratory evaluations over four runs for each ML-Accelerated variant. The Total denotes the total exploratory evaluations, while the High UCB Only column denotes the exploratory evaluations triggered by the ML model uncertainty. The standard deviation is derived from the high UCB exploratory runs, while the remaining evaluations are fixed across all runs.}
\label{tab:evals_combined}
\resizebox{0.8\textwidth}{!}{%
\begin{tabular}{l|cc|cc|cc}
\hline
 & \multicolumn{2}{c|}{$\mathrm{ML_{RF1}\text{-}MR_{ws}}$} & \multicolumn{2}{c|}{$\mathrm{ML_{RF2}\text{-}MR_{ws}}$} & \multicolumn{2}{c}{$\mathrm{ML_{GB}\text{-}MR_{ws}}$} \\
Statistic & Total & High UCB & Total & High UCB & Total & High UCB \\
\hline
Mean & 1,107,000 & 643,000 & 1,118,000 & 654,000 & 962,000 & 498,000 \\
Std  & 97,381 & 97,381 & 95,937 & 95,937 & 22,361 & 22,361 \\
Min  & 1,008,000 & 544,000 & 1,020,000 & 556,000 & 924,000 & 460,000 \\
Max  & 1,264,000 & 800,000 & 1,248,000 & 784,000 & 980,000 & 516,000 \\
\hline
\end{tabular}%
}
\end{table}

\subsection{ML Models Analysis}

\subsubsection{Metrics}
To analyze the effectiveness of the ML-Accelerated multi-resolution framework's uncertainty-driven exploration strategy, we evaluate the high uncertainty exploration decisions observed across the 4 runs. We focus on two constraints: the terminal SoC and the thermal storage median temperature $T_\text{median}$. Let $N_{unc}$ denote the total number of high uncertainty-triggered explorations. For each exploration $i$, let $y_{i}^{\text{soc}}$ and $y_{i}^{\text{temp}}$ represent the actual terminal SoC and terminal $T_\text{median}$ values, respectively. We establish uncertainty thresholds of $\theta_{\text{unc,soc}} = 0.01$ and $\theta_{\text{unc,T$_\text{median}$}} = 344.15$ K.  We define True Positives (TP) as cases where exploration revealed a significant violation as
\begin{equation}
\text{TP} = \sum_{i=1}^{N_{unc}} \mathbb{1}(y_i > \theta_{\text{unc}})
\end{equation}
and False Positives (FP) as explorations that found negligible violations as
\begin{equation}
\text{FP} = \sum_{i=1}^{N_{unc}} \mathbb{1}(y_i \leq \theta_{\text{unc}}),
\end{equation}
where $\mathbb{1}(\cdot)$ is the indicator function. The Precision (Prec) measures the fraction of explorations that discovered significant violations as
\begin{equation}
\text{Precision (\%)} = \frac{\text{TP}}{\text{TP} + \text{FP}} = \frac{100\space\text{TP}}{N_{unc}}.
\end{equation}
The Prediction Interval Coverage Probability (PICP) quantifies the calibration of uncertainty estimates by measuring how often the true value falls within the predicted upper confidence bound \cite{bazionis2021review} as
\begin{equation}
\text{PICP (\%)} = \frac{100}{N_{unc}} \sum_{i=1}^{N_{unc}} \mathbb{1}(y_i \leq u_i).
\end{equation}
Finally, we evaluate the conservativeness of the uncertainty quantification using the Normalized Mean Interval Width (NMIW). To make the width interpretable across different physical units, we express it as a percentage of the feasible operating range:
\begin{equation}
\text{NMIW (\%)} = \frac{100}{N_{unc}} \sum_{i=1}^{N_{unc}} \frac{(u_i - \hat{y}i)}{R_{\text{target}}},
\end{equation}
where $\hat{y}_i$ represents the model's predicted mean value, and $R_{\text{target}}$ denotes the physical operating range of the target variable ($R_{\text{soc}}=1$ for battery SoC and $R_{\text{temp}} = T_{\max} - T_{\min} = 20$ K for the thermal storage).

\subsubsection{Performance Analysis}

The analysis of the uncertainty-driven bypass conditions, summarized in Tables \ref{tab:unc_analysis_GB}, \ref{tab:unc_analysis_RF1}, and \ref{tab:unc_analysis_RF2}, evaluates the efficiency of the "Compute Path" logic defined in Algorithm \ref{alg:determine_target_general}. The metrics reveal a consistently exploitation-averse behavior in the triggering mechanism. The System Precision, which measures the proportion of uncertainty-triggered optimizations that properly triggered exploration, ranges between 24.7\% and 29.4\% across the three models. This indicates that in approximately 75\% of cases where the algorithm detected high uncertainty and reverted to the computationally expensive exploratory optimization. While this implies a computational overhead due to "false alarms," it confirms that the bypass condition effectively prioritizes exploration, ensuring that the fast "Predict Path" is never taken when the model suspects a potential violation.

Decomposing the triggers by objective reveals how the different physical dynamics influence the optimization logic. The strategy is notably conservative regarding the battery's SoC, with precision values consistently remaining below 7\%. This indicates that the uncertainty bounds frequently exceed the exploration threshold $\theta_{\text{unc, SoC}}$, triggering the full solver even when the battery operation should be greedy. Conversely, the thermal dynamics drive the majority of the justified computational expenditures. The thermal precision is significantly higher (averaging $\approx$27\%), suggesting that when the model flags uncertainty, there is a higher probability that the system really needs exploratory optimization for the $T_\text{median}$.

To assess the reliability of these triggers, we examine the interplay between the PICP, which measures the frequency with which the true terminal state falls within the predicted uncertainty envelope, and the NMIW, which quantifies how much of the physical operating range is consumed by that uncertainty buffer. The PICP remains high across all experiments, particularly for the RF with $\kappa=2$ model (Table \ref{tab:unc_analysis_RF2}), which achieves aggregate coverage of over 91\% for both objectives. Physically, this means that in over 90\% of cases where the model flagged uncertainty, the actual terminal state fell within the predicted bounds. However, high coverage is only valuable if the bounds are tight enough to be operationally useful; a trivial model could achieve 100\% PICP simply by predicting the entire physical range. The NMIW metric validates that the high PICP scores are achieved with meaningful bounds. For instance, for the thermal storage, the RF model with $\kappa=2$ generates significantly wider confidence bounds (NMIW $\approx 67.5\%$) compared to the $\kappa=1$ configuration ($\approx 36.3\%$). In contrast, the SoC uncertainty remains tight, consuming only $\approx 7.1\%$ of the battery's capacity while maintaining $>92\%$ coverage. Ultimately, the combination of high PICP and bounded NMIW confirms that the uncertainty-driven strategy functions as a robust net, creating necessary, physically constrained buffers rather than vague, infinite estimates. Furthermore, it incurs the cost of precautionary computations to guarantee that the system's performance rarely falls outside the predicted worst-case scenario, building confidence in the algorithm's ability to bypass the expensive optimization loop when the model is certain.

\begin{table}[H]
\centering
\caption{Uncertainty Exploration Analysis for GB. System column represents the union of violations. All metrics in \%.}
\label{tab:unc_analysis_GB}
\resizebox{0.8\textwidth}{!}{%
\begin{tabular}{c|c|ccc|ccc|c}
\hline
 & Total & \multicolumn{3}{c|}{\textbf{SoC (\%)}} & \multicolumn{3}{c|}{\textbf{T$_\text{median}$ (\%)}} & \textbf{System} \\
Run & Evals & Prec & PICP & NMIW & Prec & PICP & NMIW & Prec \\
\hline
1 & 115 & 2.6 & 81.7 & 0.07 & 20.0 & 85.2 & 47.19 & 21.7 \\
2 & 126 & 5.6 & 87.3 & 0.50 & 31.7 & 83.3 & 43.23 & 31.7 \\
3 & 129 & 4.7 & 89.1 & 0.36 & 31.8 & 80.6 & 48.54 & 31.8 \\
4 & 128 & 6.2 & 83.6 & 1.08 & 25.8 & 89.1 & 48.20 & 28.1 \\
\hline
\textbf{Avg} & 124 & \textbf{4.8} & 85.4 & 0.51 & \textbf{27.5} & 84.6 & 46.79 & \textbf{28.5} \\
\hline
\end{tabular}%
}
\end{table}

\begin{table}[H]
\centering
\caption{Uncertainty Exploration Analysis for RF1. System column represents the union of violations. All metrics in \%.}
\label{tab:unc_analysis_RF1}
\resizebox{0.8\textwidth}{!}{%
\begin{tabular}{c|c|ccc|ccc|c}
\hline
 & Total & \multicolumn{3}{c|}{\textbf{SoC (\%)}} & \multicolumn{3}{c|}{\textbf{T$_\text{median}$ (\%)}} & \textbf{System} \\
Run & Evals & Prec & PICP & NMIW & Prec & PICP & NMIW & Prec \\
\hline
1 & 139 & 4.3 & 84.2 & 1.06 & 30.2 & 79.9 & 40.27 & 32.4 \\
2 & 142 & 7.7 & 87.3 & 3.39 & 27.5 & 81.7 & 37.13 & 30.3 \\
3 & 196 & 5.6 & 89.3 & 4.75 & 21.9 & 85.7 & 33.04 & 24.0 \\
4 & 177 & 5.1 & 91.0 & 5.20 & 29.4 & 82.5 & 34.76 & 32.2 \\
\hline
\textbf{Avg} & 163 & \textbf{5.7} & 87.9 & 3.60 & \textbf{26.9} & 82.4 & 36.30 & \textbf{29.4} \\
\hline
\end{tabular}%
}
\end{table}

\begin{table}[H]
\centering
\caption{Uncertainty Exploration Analysis for RF1. System column represents the union of violations. All metrics in \%.}
\label{tab:unc_analysis_RF2}
\resizebox{0.8\textwidth}{!}{%
\begin{tabular}{c|c|ccc|ccc|c}
\hline
 & Total & \multicolumn{3}{c|}{\textbf{SoC (\%)}} & \multicolumn{3}{c|}{\textbf{T$_\text{median}$ (\%)}} & \textbf{System} \\
Run & Evals & Prec & PICP & NMIW & Prec & PICP & NMIW & Prec \\
\hline
1 & 161 & 6.2 & 90.7 & 4.68 & 24.8 & 91.9 & 69.37 & 25.5 \\
2 & 200 & 2.0 & 95.5 & 1.93 & 21.5 & 94.5 & 69.63 & 22.0 \\
3 & 136 & 9.6 & 91.9 & 9.71 & 22.8 & 90.4 & 60.84 & 24.3 \\
4 & 146 & 9.6 & 92.5 & 11.86 & 25.3 & 89.7 & 69.95 & 28.1 \\
\hline
\textbf{Avg} & 160 & \textbf{6.4} & 92.6 & 7.05 & \textbf{23.5} & 91.6 & 67.45 & \textbf{24.7} \\
\hline
\end{tabular}%
}
\end{table}

\subsection{Ablation Study}\label{subsec:ablation}

In this section, we present the results of an ablation study designed to evaluate the contributions of both the multi-resolution approach and the elite warm starting strategy. We compare four different methodologies: $\text{MR}_{ws}$, which represents the multi-resolution approach with elite warm starting; $\text{HR}$, the high-resolution baseline approach without any form of warm starting; $\text{MR}$, which denotes the multi-resolution approach lacking the elite warm starting mechanism; and $\text{HR}_{ws}$, the high-resolution baseline approach that incorporates the elite warm-starting technique. The elite warm starting mechanism implemented in $\text{HR}_{ws}$ operates by utilizing only half of the elite solutions as seeds (through random selection), aligning with the mechanics of the other approaches, while the remaining half, typically sourced from lower-resolution models, are randomly initialized. 

For the ablation study, we skipped the exploratory stage to save computational time and focus on comparing the core performance of each approach. We ran each method to control the system for 7 days and repeated each experiment 30 times to account for variability and ensure reliable results. All other hyperparameter have remained the same as in the yearly numerical experiments. 

Figure \ref{fig:ablation} illustrates the mean operating cost over a 7-day period for each of the evaluated methods. The $\text{HR}$ approach, which lacks the elite warm starting mechanism, performs the least effectively with a mean total cost of \$4,202. In comparison, the $\text{HR}_{ws}$ method, which incorporates elite warm starting, demonstrates a significant improvement, reducing the cost by 6\% (\$3,954). This result is notably comparable to that of the Rule-based controller, which recorded an operating cost of \$3,991. These results indicate that introducing warm starting allows the HR method to achieve comparable performance to the baseline rule-based controller,  clearly highlighting the effectiveness of the elite warm starting strategy. 

Additionally, the multi-resolution strategy without warm starting, denoted as $\text{MR}$, achieves a total operating cost of \$3,864 (an additional 2\% decrease), outperforming both the $\text{HR}_{ws}$ method and the Rule-based controller. This result underscores the advantages of the multi-resolution approach. The top-performing method
is the multi-resolution strategy with elite warm starting, $\text{MR}_{ws}$, which records a total cost of \$3,760 and the smallest standard deviation of \$29. This suggests a boost in performance and enhanced stability in outcomes, further reinforcing the value of integrating both elite warm starting and multi-resolution approaches within the framework.

\begin{figure}[H]
    \centering
    \includegraphics[width=0.75\linewidth]{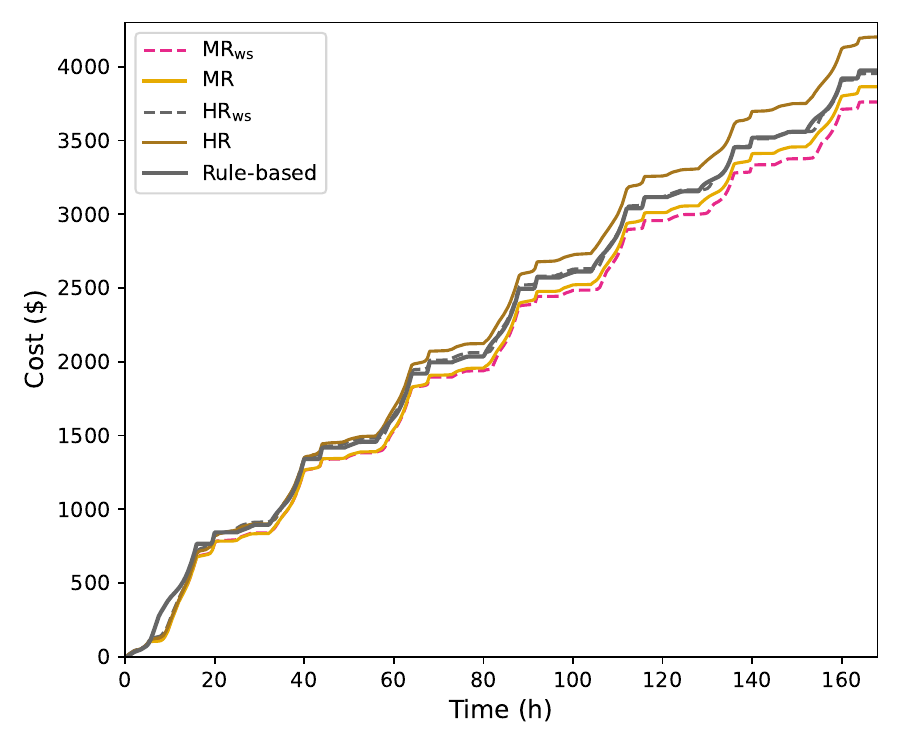}
    \caption{Mean total operating cost of the energy system over one week of all compared methods for ablation purposes. A lower operating cost at any point in the week indicates better performance.}
    \label{fig:ablation}
\end{figure}

\begin{table}[H]
\centering
\caption{Cost statistics for the methods in the ablation study.}
\label{tab:cost_statistics_full}
\begin{tabular}{lccccc}
\hline
Approach & Mean (\$) & Median (\$) & Min (\$) & Max (\$) & Std (\$) \\
\hline
$\mathrm{MR}$ & 3,864 & 3,856 & 3,808 & 4,034 & 46 \\
$\mathrm{MR_{ws}}$ & 3,760 & 3,752 & 3,715 & 3,823 & 29 \\
$\mathrm{HR}$ & 4,202 & 4,170 & 4,026 & 4,533 & 117 \\
$\mathrm{HR_{ws}}$ & 3,954 & 3,897 & 3,775 & 4,308 & 150 \\
Rule-based$^*$ & 3,991 & -- & -- & -- & -- \\
IDAES$^*$ & 2,727 & -- & -- & -- & -- \\
\hline
\multicolumn{6}{l}{$^*$Single run} \\
\end{tabular}
\end{table}

\subsection{Operating profile comparisons}\label{subsec:op-behaviour-comps}
To further assess the performance of the ML-based multi-resolution strategy, we compare the system’s operational characteristics under this control approach to the idealized results from the IDAES optimization.

\begin{figure}
    \centering
    \includegraphics[width=0.9\linewidth]{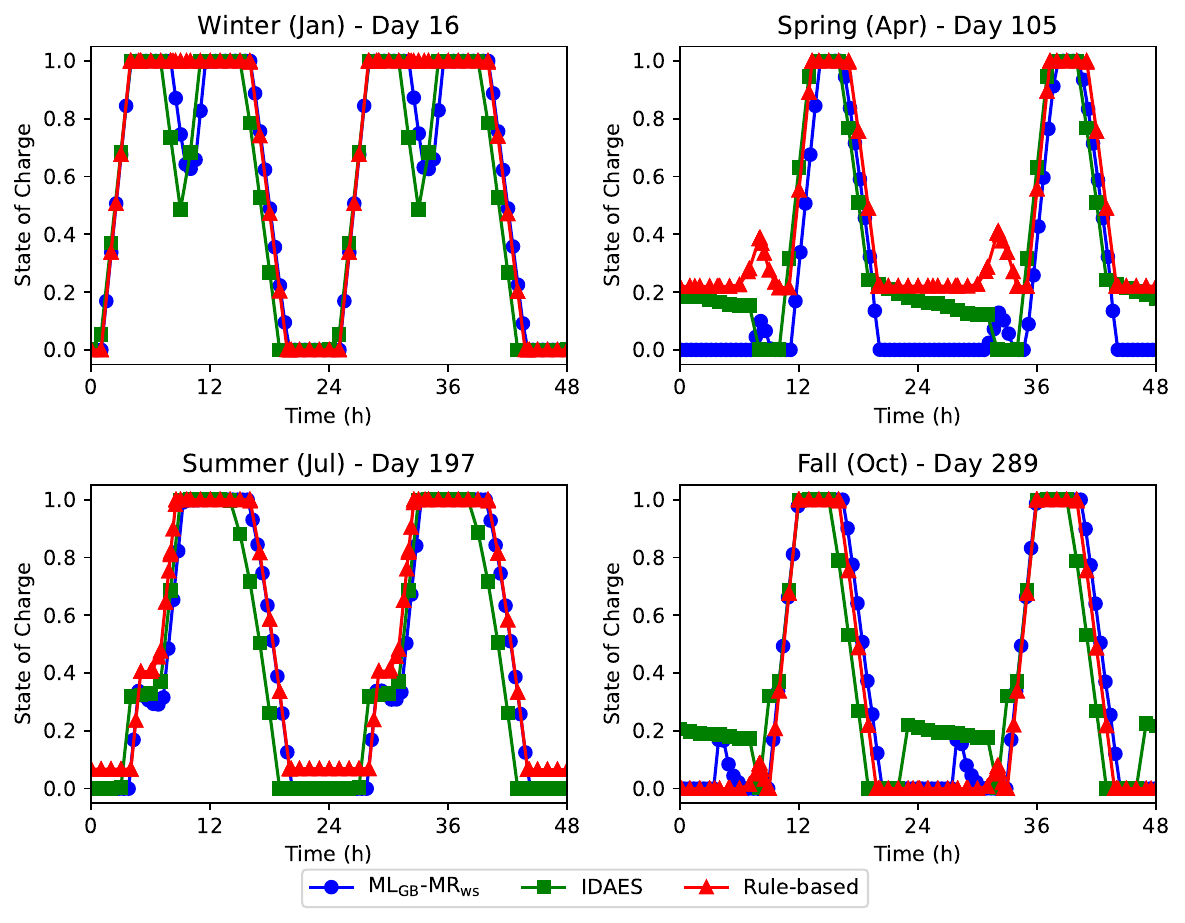}
    \caption{SOC comparison from different control strategies across seasons.}
    \label{fig:soc-comparison}
\end{figure}

Figure \ref{fig:soc-comparison} shows results for the state of charge of the battery across different seasons in the year. For comparison, we also show the result of the Rule-based control strategy. The ML-based multi-resolution strategy captures most of the trends present in the IDAES solution. This is particularly evident in the Winter case, where the ML-based controller mirrors the IDAES strategy by discharging the battery at midday, coinciding with peak PV generation. In Winter, electricity is cheapest in the morning hours (Fig. \ref{fig:elec_prices}, 0-5h), unlike in other seasons where electricity is cheapest in the middle of the day (10-15h). As a result, in Winter, all three control strategies charge the battery to full capacity early in the morning. However, the IDAES and ML-based controllers further exploit small Winter price differentials to make additional cost savings by discharging the fully charged battery at the start of plant operation (8–10 h, when electricity is slightly more expensive), and recharging a few hours later (10–12 h), when electricity is cheaper. This is overlooked by the rule-based controller, highlighting the ML approach’s ability to capture complex operational trade-offs. The ML-based controller also tracks the IDAES behavior more closely than the Rule-based approach in summer, deciding to empty the battery at the end of each day. The most significant deviation we observe in the state of charge is in Spring, where the ML-based strategy discharges the battery completely at the end of the day while the other two approaches carry over energy between days. Despite this difference however, the ML-based strategy still consistently ends up with lower daily operating costs than the Rule-based approach during this season.
A similar trend is observed when comparing the 
grid electricity purchase for the two control strategies where the ML-based controller makes similar purchase decisions to the IDAES model; further details may be found in Appendix \ref{app:operational-analysis}. For the tank temperature, the ML-based strategy was found to produce identical profiles to the rule-based controller, suggesting that the cost savings primarily stem from better battery (SOC) control.

\section{Limitations and Future Research}\label{sec: limitations}

This study demonstrated the efficacy of an ML-accelerated multi-resolution control framework applied to an industrial heating system. The primary objective was to reduce the computational burden of optimal control without sacrificing performance, achieved by learning to predict the optimal terminal states of long-horizon trajectories. By establishing this performance baseline in a controlled environment, we confirmed that ML can effectively bypass expensive optimization steps while obtaining the best operating cost of the evaluated controllers.

To further enhance the algorithm's efficiency, future work should focus on refining the uncertainty quantification that drives the exploration trigger. While our analysis of the GB and RF models demonstrated high probabilistic coverage (PICP), the low system precision indicates that the trigger is often overly conservative. Although the GB model improves upon the heuristic variance of the RF by using quantile loss, it remains dependent on a fixed, manually selected quantile parameter ($\alpha$) to define the exploration margin. Replacing these with more rigorous UQ frameworks, such as Conformal Prediction, could yield tighter, statistically guaranteed bounds. This would improve the precision of the trigger and reducing the frequency of unnecessary computational expenses.

A parallel enhancement to the triggering logic is to evolve the static uncertainty threshold, $\theta_{unc}$, into a dynamic, state-dependent parameter. The current use of a fixed threshold presents a significant limitation: it requires manual calibration and assumes a uniform tolerance for uncertainty across all operating regimes. A dynamic threshold would address this by allowing the controller to autonomously adjust its sensitivity based on the system's proximity to constraints or historical volatility. This would eliminate the need for manual tuning and ensure the algorithm adapts its exploration behavior as it encounters new operating conditions.

Finally, while this study utilized deterministic price forecasts to provide a clear assessment of the algorithm's core mechanics, real-world deployment must account for forecast errors. The current model is reactive to a single predicted future rather than being robust to a range of plausible scenarios. A significant avenue for future research is to integrate stochastic or ensemble-based forecasts.
\section{Conclusion}
In this work, we sought to understand the gap between the optimization and verification models used in industrial energy systems design via a limit of performance analysis. To achieve this, we propose an online, ML–accelerated multi-resolution optimization framework that estimates architecture-specific upper bounds on achievable performance while minimizing the number of expensive high-fidelity model evaluations. Our ML-guided controller adaptively schedules the optimization resolution based on predictive uncertainty and warm-starts the high-resolution optimization solver using elite low-resolution solutions. We demonstrate our approach on the pilot case of heating an industrial process. Our results show that the multi-resolution strategy improves both controller performance and optimization tractability, yielding operating cost reductions of up to 24\% relative to direct high-resolution optimization and up to 10.5\% relative to rule-based control. We show that integrating online, uncertainty-based ML strategies for decision-making on design space exploration can provide significant computational savings without compromising solution quality. Gradient Boosting in particular shows significant potential for our application due to its distribution-free approach to uncertainty quantification. An analysis of the proposed control strategies shows that the ML-based controller mimics the behavior of the idealized IDAES-based control strategy, suggesting
that the ML-based approach converges to a high-quality solution. Our proposed approach of combining multi-resolution optimization and ML makes high-fidelity verification tractable and computationally efficient for approximating the limit-of-performance, providing a concrete way to quantify achievable energy-system operating performance prior to industrial deployment.

\section*{Declaration of Competing Interest}
The authors declare that they have no known competing financial interests or personal relationships that could have appeared to influence the work reported in this paper.

\section*{Data Availability}
The code needed to reproduce the study can be found on the following repository: \url{https://github.com/lbl-srg/pbd-industrial-limit-of-performance}. 

\section*{Acknowledgments}
This work was supported by the Laboratory Directed Research and Development Program of Lawrence Berkeley National Laboratory under U.S. Department of Energy Contract No. DE-AC02-05CH11231.

This research used the Lawrencium computational cluster resource provided by the IT Division at the Lawrence Berkeley National Laboratory (Supported by the Director, Office of Science, Office of Basic Energy Sciences, of the U.S. Department of Energy under Contract No. DE-AC02-05CH11231).

\newpage
\appendix
\counterwithin{figure}{section}
\counterwithin{table}{section}
\renewcommand{\thefigure}{\thesection.\arabic{figure}}
\renewcommand{\thetable}{\thesection.\arabic{table}}
\section{IDAES Model Equations}\label{app:process-models}

\subsection{Costing}
\subsubsection{Capital costs}
The capital cost of the system is computed as a sum of the costs of the fixed and variable capital costs of the individual components in the overall system architecture,
\begin{equation*}
C_{\mathrm{cap}}=\sum_{k} \tilde{C}_{\text{fixed},k}+\tilde{C}_{\text{var},k} \, B_{k},
\end{equation*}
where $k$ represents the generation and storage technologies available (PV, gas boiler, heat pump, storage tank, battery),$\tilde{C}_{\text{fixed},k}$ are the fixed costs of the components $k$, $\tilde{C}_{\text{var},k}$ are the variable capital costs, and $B_{k}$ are the sizing parameters for the different units.

Table~\ref{tab:cap-cost-tables} shows the fixed and variable capital costs for each unit, along with the sizing parameters.

\begin{table}[H]
\caption{Sizing and unit cost parameters for energy system}\label{tab:cap-cost-tables}
\centering{}%
\begin{tabular}{ll>{\raggedleft}m{2.5cm}>{\raggedleft}m{3.5cm}c}
\toprule 
Component & Sizing parameter $B_{k}$ & Fixed cost $\ensuremath{\tilde{C}_{\text{fixed},k}}$ (USD) & Variable capital cost $\ensuremath{\tilde{C}_{\text{var},k}}$ & Ref.\tabularnewline
\midrule
PV & Capacity - power (kW) & 0 & 3500 $\unitfrac{USD}{kW}$ & \cite{southland} \tabularnewline
Gas boiler & Capacity - power (kW) & 79670 & 367 $\unitfrac{USD}{kW}$ & \cite{southland}\tabularnewline
Heat pump & Capacity - power (kW) & 0 & 2055 $\unitfrac{USD}{kW}$ & \cite{southland} \tabularnewline
Battery & Capacity - energy (kWh) & 0 & 757 $\unitfrac{USD}{kWh}$ & \cite{pvprice} \tabularnewline
Hot water tank & Capacity - volume (m\textsuperscript{3}) & 1520 & 4845 $\unitfrac{USD}{m^{3}}$ & \cite{southland} \tabularnewline
\bottomrule
\end{tabular}
\end{table}

\subsubsection{Annual operating costs}
The annual operating cost is computed as the sum of the hourly costs of electricity purchase from the grid and gas purchase for the boiler over the entire year,
\begin{equation*}
C_{\mathrm{op}}=\sum_{t=1}^{8760} \tilde{C}_{\text{op,elec}} \, E_{\mathrm{grid}}(t, \mathbf{x}) + \tilde{C}_{\text{op,gas}} \,  E_{\mathrm{gas}}(t, \mathbf{x}),
\end{equation*}
where $E_{\mathrm{grid}}(t, \mathbf{x})$ and $E_{\mathrm{gas}}(t, \mathbf{x})$ are the hourly energies from the grid and gas purchases, respectively. For our analysis, we use a fixed gas price of $\unitfrac[0.039]{USD}{kWh}$. 

\subsection{Process Models}
\subsubsection{Photovoltaics}
The PV model calculates the electricity output of the PV system $P_{PV}$ as a function of the solar irradiation $G$ and the installed PV area $A$ as
\begin{equation*}
    P_{PV} = \eta_{PV} \, A \, G,
\end{equation*}
where $\eta_{PV} = 17.1\%$ is the PV efficiency.

\subsubsection{Gas boiler}
The gas boiler computes the power output from the boiler $\dot{Q}_{boiler}$ (kW) as a function of the volumetric gas flow rate to the boiler $\upsilon_{gas}$ as
\begin{equation*}
    \dot{Q}_{boiler} = \eta_{boiler} \, \upsilon_{gas} \, \mathrm{HHV},
\end{equation*}
where $\mathrm{HHV}= 35396 kJ/m\textsuperscript{3}$ is the higher heating value of natural gas. The energy generated from the boiler raises the temperature water based on the heat transfer equation as
\begin{equation*}
\dot{Q}_{boiler} = m_{water} \, C_{p} \, \Delta T_{water} ,
\end{equation*}
where $m_{water}$ is the mass flow rate into the boiler (kg/s).

\subsubsection{Heat pump}
In addition to the heat transfer equations on the evaporator and condenser sides of the heat pump, the heat pump model includes the overall energy balance
\begin{equation*}
\dot{Q}_{condenser} + P_{HP} = \dot{Q}_{evaporator},
\end{equation*}
where $\dot{Q}_{condenser}$ and $\dot{Q}_{evaporator}$ represent the energy rates on the condenser and evaporator sides, respectively, and $P_{HP}$ is the electricity into the heat pump. The performance equation of the heat pump is based on a fixed coefficient of performance $\eta_{COP}$,
\begin{equation*}
\eta_{COP} P_{HP} = \dot{Q}_{evaporator}
\end{equation*}

\subsubsection{Battery}
The battery module is based on the standard dynamic battery model that represents the rate of change of energy within the battery as a function of the instantaneous energy rates into and out of the battery while accounting for charging and discharging losses. For the battery model, we constrain the maximum rate of charge (i.e., battery energy to power capacity) to be greater than three hours. 

Further details about battery modeling may be found in~\citet{AMUSAT2018379}. 

\subsubsection{Hot water tank}
The hot water tank is modeled as a cylindrical storage tank with perfect mixing assumptions, hence the temperature within the tank is considered to be uniform with no stratification. The model consists of dynamic energy balances for the storage tank, and accounts for thermal losses from the storage tank to the surroundings. The thermal loss from the tank $\dot{Q}_{loss}$ is computed as a function of the tank state of charge as
\begin{equation*}
\dot{Q}_{loss} = U_{tank} \, A_{tank} \,(T_{tank} - T_{ambient}) \, \mathrm{SoC}_{tank},
\end{equation*}
where $U_{tank}$, $A_{tank}$, $T_{tank}$, and $T_{ambient}$ denote the overall heat transfer coefficient, tank surface area, tank temperature, and ambient temperature, respectively, and $\mathrm{SoC}_{tank}$ represents the tank state of charge, defined as the fraction of stored energy relative to the installed capacity.

\section{Modelica Modeling}\label{app:modelica-models}

\subsection{Heat production}

Heat is provided using a heat pump or a gas boiler. 

The heat pump model is \textit{Buildings.Fluid.HeatPumps.Carnot\_TCon}.
It takes as a control input the condenser leaving temperature and
it computes the coefficient of performance at off-design conditions based
on a Carnot efficiency.
The coefficient of performance at design conditions is $COP_{0}=3$. The electric power consumption of the heat pump compressor is
\begin{equation*}
	\label{eq:cop}
	P_{HP} = \frac{\dot Q_{con}}{COP},
\end{equation*}
where $COP$ is the coefficient of performance at the current operating temperatures and $\dot Q_{con}$ is the condenser heat flow rate.

The gas boiler model is \textit{Buildings.Fluid.Boilers.BoilerPolynomial}. It computes the gas consumption as
\begin{equation*}
	\dot Q_{gas} = \frac{\dot Q_{use}}{\eta_{boiler}},
\end{equation*}
where
$\dot Q_{use}$ is the useful heat and
$\eta_{boiler}=0.9$ is the efficiency.

\subsection{Energy storage}

There is a battery for electric storage, and a hot water tank for thermal storage.

The battery model is \textit{Buildings.Electrical.DC.Storage.Battery}, which uses charging and discharging efficiency of $\eta = 0.95$

The thermal storage model is \textit{Buildings.Fluid.Storage.StratifiedEnhanced}.
It models a stratified storage tank with heat loss to
the environment.
The tank also exposes the fluid temperature at various heights,
which are used by the rule-based controller described in Section~\ref{sec:rule_based_control}.

\subsection{Electricity production}

The electricity is either imported from the grid or locally produced via PV. 

The PV production model is \textit{Buildings.Electrical.DC.Sources.PVSimple}.
It computes the efficiency as
\begin{equation*}
	\label{eq:etapv}
	\eta_{PV} = f_{act} \, \eta_{conv} \, (1 - \eta_{loss}) = 0.171,
\end{equation*}
where $f_{act} = 0.9$ is the fraction of the active solar cell surface area, $\eta_{conv} = 0.2$ is the initial system conversion efficiency, considering PV panels, wiring, and inverters \cite{nrel_eff}, and $\eta_{loss} = 0.05$ is the conversion loss after 10 years \cite{JordanKurtz2013}.
Similar to the IDAES model, the produced electricity is
\begin{equation*}
	\label{eq:Ppv}
	P_{PV} = \eta_{PV} \, A \, G,
\end{equation*}
where $A$ is the collector area and $G$ is the total solar irradiation onto the PV.

\subsubsection{Mass flow rates} \label{sec:m_flow}

In the rule-based controls model, the mass flow rates of the pumps for the heat pump condenser, the gas boiler and the heat exchanger are set to
\begin{equation*} \label{eq:m_flow}
	\dot{m} = \frac{\dot Q_{0}}{c_p \, \Delta T_{0}},
\end{equation*}
where
$\dot Q_{0}$ is the nominal heat flow rate,
$c_p$ is the specific heat capacity of water, and
$\Delta T_{0} = 20 \, \mathrm{K}$ is the nominal temperature difference between the inlet and the outlet.

For the pump on the evaporative side of the heat pump,
the mass flow rate is
\begin{equation*} \label{eq:m_flow_evap}
	\dot m_{eva} = \frac{\dot Q_{eva}}{c_p \, \Delta T_{eva}},
\end{equation*}
where
$\dot Q_{eva}$ is the evaporator heat flow rate,
$\Delta T_{eva} = 4K$ is the set point of the temperature difference between the evaporator inlet and the outlet temperature.

In model that is used for the limit-of-performance analysis, the mass flow rates of the pump for the condenser and the gas boiler are provided as input from the optimization.
\section{Simulation data for Pilot Case Study}\label{app:simulation-data}

\subsection{Electricity prices}
Figure \ref{fig:elec_prices} shows the used hourly electricity price profiles. The prices are based on the fictitious, prototype research price profiles obtained from the California Load Flexibility Research and Development Hub through the CEC’s Market Informed Demand Automation Server\cite{calflexhub}.  

\begin{figure}[H]
    \centering
    \includegraphics[width=0.9\linewidth]{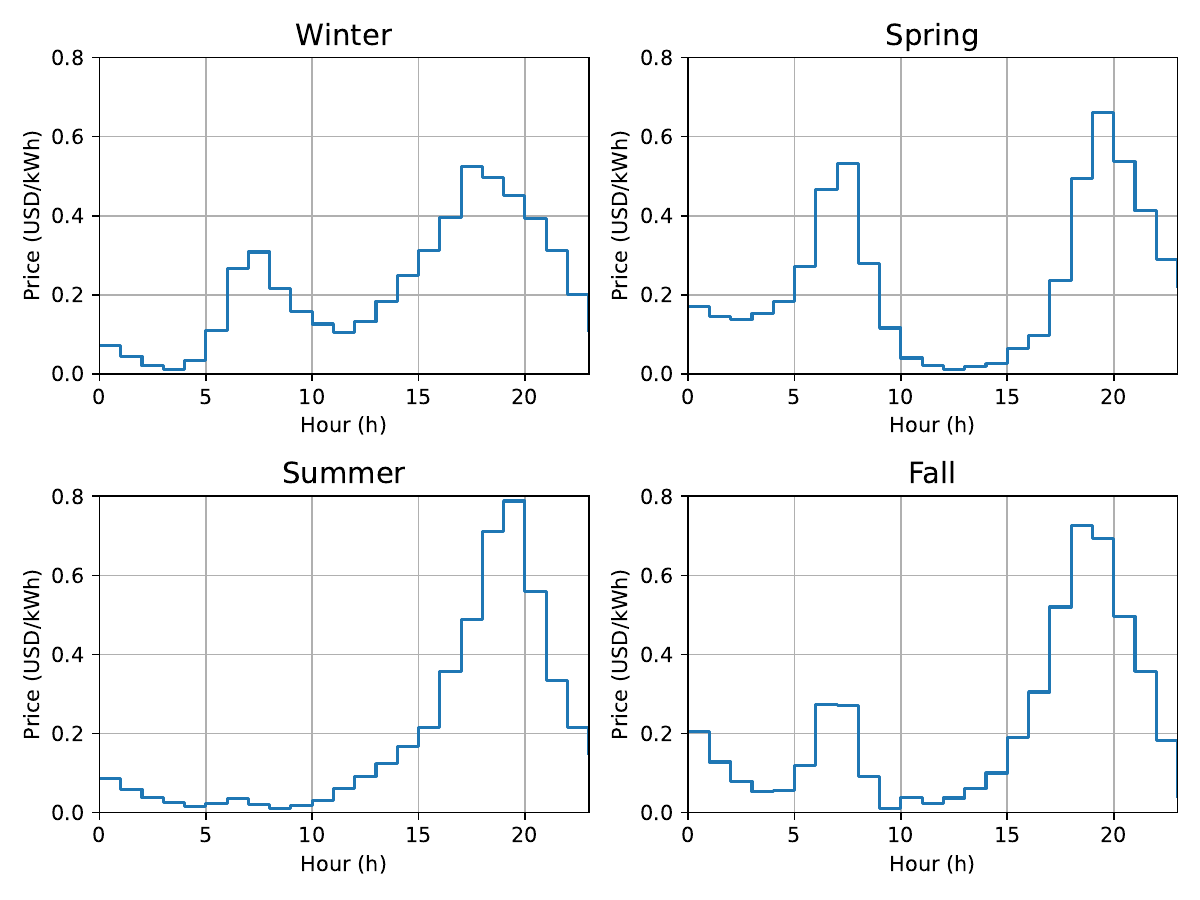}
    \caption{Hourly electricity prices per season.}
    \label{fig:elec_prices}
\end{figure}

\subsection{Model parameters for IDAES architectural optimization}

Table \ref{tab:IDAES-Model-Parameters.} summarizes the parameters used for the different components in the architectural optimization step.

\begin{table}[h]
\caption{IDAES Model Parameters.\label{tab:IDAES-Model-Parameters.}}
\begin{centering}
\begin{tabular}{clr}
\hline
Process Model & Parameter & Value\tabularnewline
\hline
Photovoltaics & PV efficiency, $\eta_{PV}$ & 17.1\%\tabularnewline
\hline
Heat Pump & Coefficient of performance, $\eta_{COP}$ & 3.0\tabularnewline
 & Hot side outlet temperature & $92^{\text{o}}C$\tabularnewline
\hline
Gas boiler & Boiler efficiency, $\eta_{boiler}$ & 0.90\tabularnewline
 & Natural gas heating value, HHV & $\unitfrac[35,396]{kJ}{m^{3}}$\tabularnewline
 & Outlet temperature & $92^{\text{o}}C$\tabularnewline
\hline
Battery & Discharging efficiency, $\eta_{discharge}$ & 0.95\tabularnewline
 & Charging efficiency, $\eta_{charge}$ & 0.95\tabularnewline
\hline
 & Ambient temperature, $T_{ambient}$ & $20^{\text{o}}C$\tabularnewline
 & Overall heat transfer coefficient, $U_{tank}$ & $0.4\unitfrac{W}{m^{2}K}$\tabularnewline
 & Minimum allowable water temperature, $T_{tank}^{min}$ & $70^{\text{o}}C$\tabularnewline
 & Maximum allowable water temperature, $T^{max}_{tank}$ & $90^{\text{o}}C$\tabularnewline
 & Tank height-to-diameter ratio & 1\tabularnewline
\hline
Load & Water outlet temperature & $70^{\text{o}}C$\tabularnewline
\hline
\end{tabular}
\par\end{centering}
\end{table}
\section{Generation and Storage Capacities for Pareto-Optimal Designs}\label{app:pareto-analysis}

Figure \ref{fig:installed-capacities} shows installed heat generation and energy storage capacities for the Pareto-optimal designs. 

\begin{figure}[H]
	\centering
	\begin{subfigure}[b]{0.49\textwidth}
		\centering
		\includegraphics[width=\linewidth]{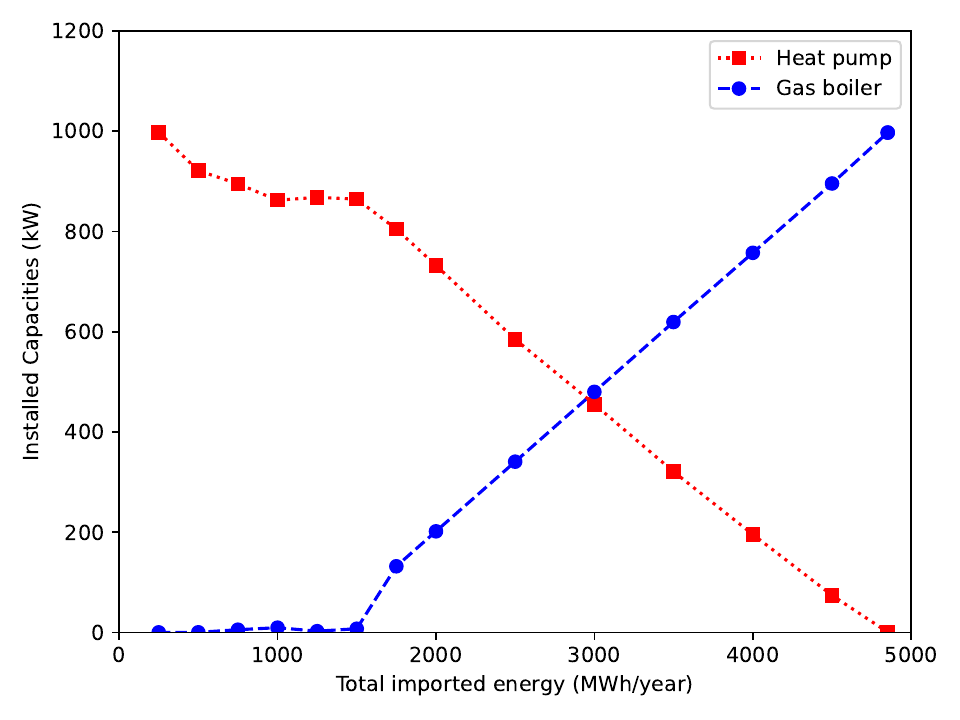}
		\caption{Heat generation}
		\label{fig:pareto-gen}
	\end{subfigure}
	\hfill 
	\begin{subfigure}[b]{0.49\textwidth}
		\centering
		\includegraphics[width=\linewidth]{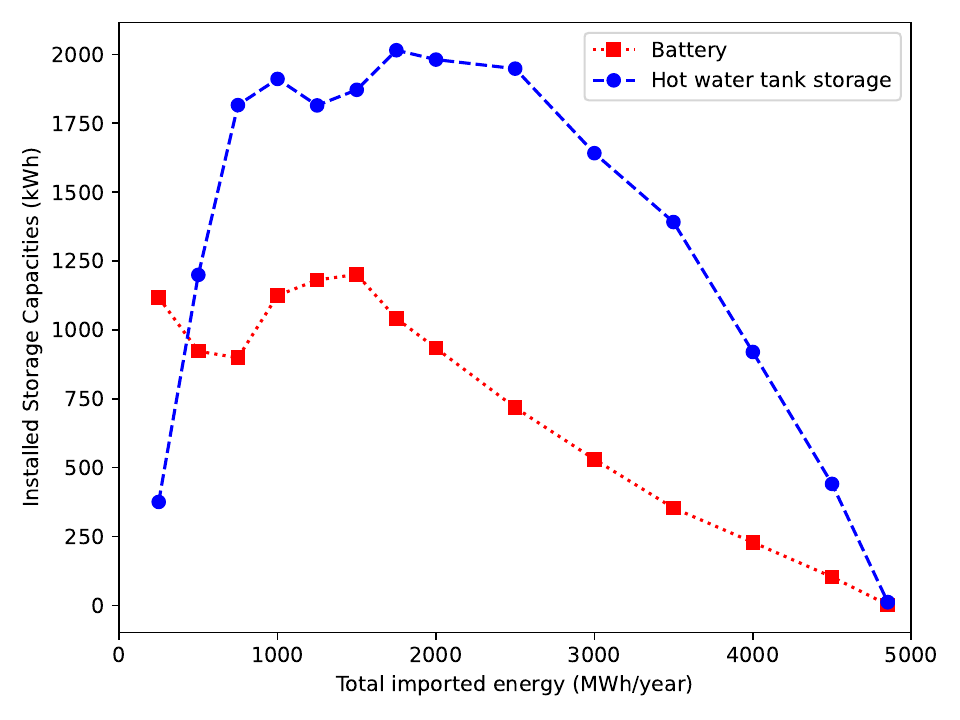}
		\caption{Storage}
		\label{fig:pareto-storage}
	\end{subfigure}
	\caption{Installed capacities of generation and storage technologies for Pareto-optimal designs}
	\label{fig:installed-capacities}
\end{figure}

\section{Comparison of Control Stategies}\label{app:operational-analysis}



Figure \ref{fig:grid-comparison} compares amount of electricity purchased from the grid over a 48-hour period in each season. Generally, the two profiles match quite well, with the ML-based controller buying electricity at the same times and at roughly the same levels as the idealized control. The only notable difference occurs in Fall, where the IDAES strategy buys electricity at end of the day to charge the tank, while the ML controller decides to buy at early morning.

\begin{figure}[H]
    \centering
    \includegraphics[width=0.95\linewidth]{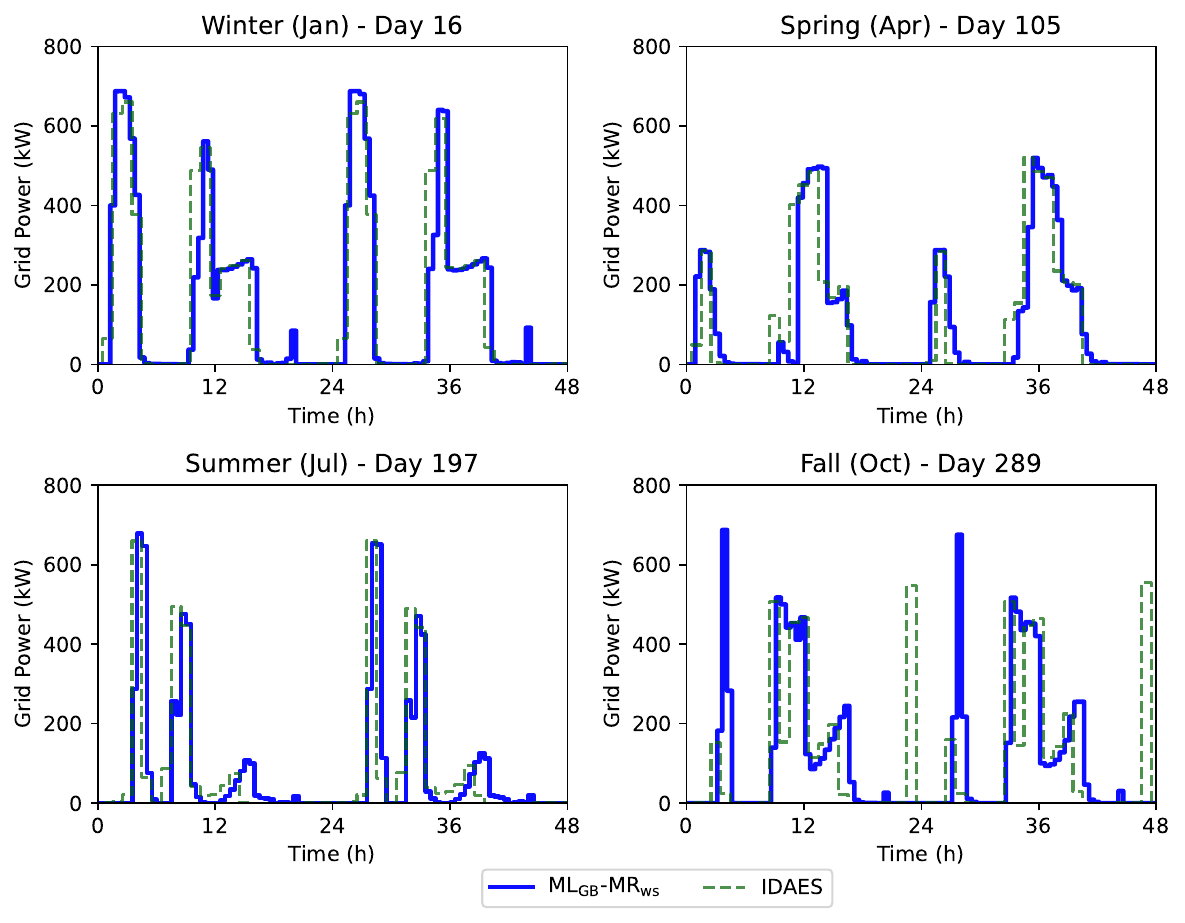}
    \caption{Electricity purchase from grid: IDAES vs. ML-based approaches.}
    \label{fig:grid-comparison}
\end{figure}
\newpage
\bibliographystyle{unsrtnat}
\bibliography{references}

@inproceedings{MattssonEtAl1999,
	Address = {Phoenix, AZ},
	Author = {S. E. Mattsson and M. Otter and H. Elmqvist},
	Booktitle = {38th IEEE Conference on Decision and Control},
	Month = dec,
	Organization = {IEEE},
	Pages = {3502--3507},
	Title = {Modelica Hybrid Modeling and Efficient Simulation},
	Url = {http://www.modelica.org/publications/papers/ModelicaCDC99.pdf},
	Year = {1999}}

@TechReport{WilcoxMarion2008,
  author = 	 {S. Wilcox and W. Marion},
  title = 	 {Users Manual for {TMY3} Data Sets},
  institution =  {NREL},
  year = 	 {2008},
  number = 	 {NREL/TP-581-43156},
  address = 	 {Golden, CO},
  month = 	 may
}

@Article{WetterZuoNouiduiPang2014,
  author = 	 {Michael Wetter and Wangda Zuo and Thierry S. Nouidui and Xiufeng Pang},
  title = 	 {Modelica {Buildings} library},
  journal = 	 {Journal of Building Performance Simulation},
  volume  =      {7},
  number  =      {4},
  pages   =      {253--270},
  year = 	 {2014},
  doi  =         {10.1080/19401493.2013.765506},
  url = "https://doi.org/10.1080/19401493.2013.765506"
}

@inproceedings{MattssonElmqvist1997:1,
	Address = {Gent, Belgium},
	Author = {Sven Erik Mattsson and Hilding Elmqvist},
	Booktitle = {7th IFAC Symposium on Computer Aided Control Systems Design},
	Editor = {L. Boullart and M. Loccufier and Sven Erik Mattsson},
	Month = apr,
        Pages = {1--5},
	Title = {Modelica -- {An} international effort to design the next generation modeling language},
	Url = {http://www.modelica.org/publications/papers/CACSD97Modelica.pdf},
	Year = {1997}}

@book{Pol97:1,
	Author = {Elijah Polak},
	Publisher = {Springer Verlag},
        Location = {New York},
	Series = {Applied Mathematical Sciences},
	Title = {Optimization, Algorithms and Consistent Approximations},
	Volume = {124},
	Year = {1997}}

@article{Sangiovanni2007:1,
    author = {Alberto Sangiovanni-Vincentelli},
    title = {Quo Vadis, {SLD}? Reasoning About the Trends and
              Challenges of System Level Design},
    journal = {Proc. of the IEEE},
    volume = {95},
    number = {3},
    pages = {467-506},
    month = {March},
    year = {2007},
    URL = {http://www.gigascale.org/pubs/1023.html},
    doi = "10.1109/JPROC.2006.890107"
}

@article{SulzerWetterMutschlerSangiovanni2023,
  author = {Matthias Sulzer and Michael Wetter and Robin Mutschler and Alberto Sangiovanni-Vincentelli},
  title = {Platform-based design for energy systems},
  journal = {Applied Energy},
  volume = {352},
  pages = {121955},
  year = {2023},
  issn = {0306-2619},
  doi = {10.1016/j.apenergy.2023.121955},
  url = {https://doi.org/10.1016/j.apenergy.2023.121955}
}

@Article{WetterSulzer2024,
author = {Michael Wetter and Matthias Sulzer},
title = {A call to action for building energy system modelling in the age of decarbonization},
journal = {Journal of Building Performance Simulation},
volume = {17},
number = {3},
pages = {383--393},
year = {2024},
publisher = {Taylor \& Francis},
doi = {10.1080/19401493.2023.2285824},
url = {https://doi.org/10.1080/19401493.2023.2285824}
}

@article{zhang2009jade,
  title={JADE: adaptive differential evolution with optional external archive},
  author={Zhang, Jingqiao and Sanderson, Arthur C},
  journal={IEEE Transactions on evolutionary computation},
  volume={13},
  number={5},
  pages={945--958},
  year={2009},
  publisher={IEEE}
}

@article{pedregosa2011scikit,
  title={Scikit-learn: Machine learning in Python},
  author={Pedregosa, Fabian and Varoquaux, Ga{\"e}l and Gramfort, Alexandre and Michel, Vincent and Thirion, Bertrand and Grisel, Olivier and Blondel, Mathieu and Prettenhofer, Peter and Weiss, Ron and Dubourg, Vincent and others},
  journal={the Journal of machine Learning research},
  volume={12},
  pages={2825--2830},
  year={2011},
  publisher={JMLR. org}
}

@article{breiman2001random,
  title={Random forests},
  author={Breiman, Leo},
  journal={Machine learning},
  volume={45},
  number={1},
  pages={5--32},
  year={2001},
  publisher={Springer}
}

@article{bazionis2021review,
  title={Review of deterministic and probabilistic wind power forecasting: Models, methods, and future research},
  author={Bazionis, Ioannis K and Georgilakis, Pavlos S},
  journal={Electricity},
  volume={2},
  number={1},
  pages={13--47},
  year={2021},
  publisher={MDPI}
}

@article{lee2021,
author = {Lee, Andrew and Ghouse, Jaffer H. and Eslick, John C. and Laird, Carl D. and Siirola, John D. and Zamarripa, Miguel A. and Gunter, Dan and Shinn, John H. and Dowling, Alexander W. and Bhattacharyya, Debangsu and Biegler, Lorenz T. and Burgard, Anthony P. and Miller, David C.},
title = {The IDAES process modeling framework and model library—Flexibility for process simulation and optimization},
journal = {Journal of Advanced Manufacturing and Processing},
volume = {3},
number = {3},
pages = {e10095},
doi = {https://doi.org/10.1002/amp2.10095},
url = {https://aiche.onlinelibrary.wiley.com/doi/abs/10.1002/amp2.10095},
eprint = {https://aiche.onlinelibrary.wiley.com/doi/pdf/10.1002/amp2.10095},
year = {2021}
}

@inproceedings{rawlings2022multiperiod,
  title={Multiperiod Generalized Disjunctive Programming Optimization in Idaes: Simultaneous Design and Operation of an Integrated Energy System},
  author={Rawlings, Edna and Ghouse, Jaffer and Susarla, Naresh and Siirola, John and Miller, David},
  booktitle={2022 AIChE Annual Meeting},
  year={2022},
  organization={AIChE}
}

@article{rao2024,
author = {Rao, Akshay K. and Atia, Adam A. and Knueven, Bernard and Mauter, Meagan S.},
title = {Optimizing Desalination Operations for Energy Flexibility},
journal = {ACS Sustainable Chemistry \& Engineering},
volume = {12},
number = {42},
pages = {15696-15704},
year = {2024},
doi = {10.1021/acssuschemeng.4c06353},
URL = {https://doi.org/10.1021/acssuschemeng.4c06353},
eprint = {https://doi.org/10.1021/acssuschemeng.4c06353}
}

@article{paszke2019pytorch,
  title={Pytorch: An imperative style, high-performance deep learning library},
  author={Paszke, Adam and Gross, Sam and Massa, Francisco and Lerer, Adam and Bradbury, James and Chanan, Gregory and Killeen, Trevor and Lin, Zeming and Gimelshein, Natalia and Antiga, Luca and others},
  journal={Advances in neural information processing systems},
  volume={32},
  year={2019}
}

@Report{pvprice,
  author      = {Bolinger, Mark and Seel,Joachim and Mulvaney Kemp, Julie and Warner, Cody and Katta, Anjali and Robson, Dana},
  date        = {2023-10},
  institution = {Lawrence Berkeley National Laboratory},
  title       = {Utility-Scale Solar, 2023 Edition},
}

@misc{southland,
  author = {{Southland industries (personal communication)}},
}

@article{AMUSAT2018379,
title = {Optimal design of hybrid energy systems incorporating stochastic renewable resources fluctuations},
journal = {Journal of Energy Storage},
volume = {15},
pages = {379-399},
year = {2018},
issn = {2352-152X},
doi = {https://doi.org/10.1016/j.est.2017.12.003},
url = {https://www.sciencedirect.com/science/article/pii/S2352152X17303365},
author = {Oluwamayowa O. Amusat and Paul R. Shearing and Eric S. Fraga},
}

@article{wachter2006,
  title={On the implementation of an interior-point filter line-search algorithm for large-scale nonlinear programming},
  author={W{\"a}chter, Andreas and Biegler, Lorenz T},
  journal={Mathematical programming},
  volume={106},
  number={1},
  pages={25--57},
  year={2006},
  publisher={Springer}
}

@Online{calflexhub,
  author = {Prakash, Anand and Paul, Lazlo},
  title  = {CalFlexHub - California Load Flexibility Research and Development Hub},
  url    = {https://github.com/LBNL-ETA/CalFlexHub},
}

@article{abdulla2017multi,
  title={Multi-resolution dynamic programming for the receding horizon control of energy storage},
  author={Abdulla, Khalid and De Hoog, Julian and Steer, Kent and Wirth, Andrew and Halgamuge, Saman},
  journal={IEEE Transactions on Sustainable Energy},
  volume={10},
  number={1},
  pages={333--343},
  year={2017},
  publisher={IEEE}
}

@article{jain2008trajectory,
  title={Trajectory optimization using multiresolution techniques},
  author={Jain, Sachin and Tsiotras, Panagiotis},
  journal={Journal of Guidance, Control, and Dynamics},
  volume={31},
  number={5},
  pages={1424--1436},
  year={2008}
}

@article{zhang2023multi,
  title={Multi-resolution based PID controller for frequency regulation of a hybrid power system with multiple interconnected systems},
  author={Zhang, Peng and Daraz, Amil and Malik, Suheel Abdullah and Sun, Chao and Basit, Abdul and Zhang, Guoqiang},
  journal={Frontiers in Energy Research},
  volume={10},
  pages={1109063},
  year={2023},
  publisher={Frontiers Media SA}
}

@article{zou2018wavelet,
  title={Wavelet multi-resolution approximation for multiobjective optimal control},
  author={Zou, Wen and Zhang, Qingbin and Gao, Qingyu and Feng, Zhiwei},
  journal={Plos one},
  volume={13},
  number={8},
  pages={e0201514},
  year={2018},
  publisher={Public Library of Science San Francisco, CA USA}
}

@article{perera2019machine,
  title={Machine learning methods to assist energy system optimization},
  author={Perera, ATD and Wickramasinghe, PU and Nik, Vahid M and Scartezzini, Jean-Louis},
  journal={Applied energy},
  volume={243},
  pages={191--205},
  year={2019},
  publisher={Elsevier}
}

@article{bre2020efficient,
  title={An efficient metamodel-based method to carry out multi-objective building performance optimizations},
  author={Bre, Facundo and Roman, Nadia and Fachinotti, V{\'\i}ctor D},
  journal={Energy and buildings},
  volume={206},
  pages={109576},
  year={2020},
  publisher={Elsevier}
}

@article{prina2024machine,
  title={Machine learning as a surrogate model for EnergyPLAN: Speeding up energy system optimization at the country level},
  author={Prina, Matteo Giacomo and Dallapiccola, Mattia and Moser, David and Sparber, Wolfram},
  journal={Energy},
  volume={307},
  pages={132735},
  year={2024},
  publisher={Elsevier}
}

@article{stoffel2023safe,
  title={Safe operation of online learning data driven model predictive control of building energy systems},
  author={Stoffel, Phillip and Henkel, Patrick and R{\"a}tz, Martin and K{\"u}mpel, Alexander and M{\"u}ller, Dirk},
  journal={Energy and AI},
  volume={14},
  pages={100296},
  year={2023},
  publisher={Elsevier}
}

@article{ratz2024identifying,
  title={Identifying the validity domain of machine learning models in building energy systems},
  author={R{\"a}tz, Martin and Henkel, Patrick and Stoffel, Phillip and Streblow, Rita and M{\"u}ller, Dirk},
  journal={Energy and AI},
  volume={15},
  pages={100324},
  year={2024},
  publisher={Elsevier}
}

@article{grbcic2024efficient,
  title={Efficient inverse design optimization through multi-fidelity simulations, machine learning, and boundary refinement strategies},
  author={Grbcic, Luka and M{\"u}ller, Juliane and de Jong, Wibe Albert},
  journal={Engineering with Computers},
  volume={40},
  number={6},
  pages={4081--4108},
  year={2024},
  publisher={Springer}
}

@article{liang2024survey,
  title={A survey of surrogate-assisted evolutionary algorithms for expensive optimization},
  author={Liang, Jing and Lou, Yahang and Yu, Mingyuan and Bi, Ying and Yu, Kunjie},
  journal={Journal of Membrane Computing},
  pages={1--20},
  year={2024},
  publisher={Springer}
}

@article{favaro2024multi,
  title={Multi-fidelity optimization for the day-ahead scheduling of Pumped Hydro Energy Storage},
  author={Favaro, Pietro and Gobert, Maxime and Toubeau, Jean-Fran{\c{c}}ois},
  journal={Journal of Energy Storage},
  volume={103},
  pages={114096},
  year={2024},
  publisher={Elsevier}
}

@article{liu2023surrogate,
  title={Surrogate-assisted many-objective optimization of building energy management},
  author={Liu, Qiqi and Lanfermann, Felix and Rodemann, Tobias and Olhofer, Markus and Jin, Yaochu},
  journal={IEEE Computational Intelligence Magazine},
  volume={18},
  number={4},
  pages={14--28},
  year={2023},
  publisher={IEEE}
}

@article{li2023deep,
  title={Deep reinforcement learning for smart grid operations: Algorithms, applications, and prospects},
  author={Li, Yuanzheng and Yu, Chaofan and Shahidehpour, Mohammad and Yang, Tao and Zeng, Zhigang and Chai, Tianyou},
  journal={Proceedings of the IEEE},
  volume={111},
  number={9},
  pages={1055--1096},
  year={2023},
  publisher={IEEE}
}

@article{michailidis2025reinforcement,
  title={Reinforcement learning for optimizing renewable energy utilization in buildings: A review on applications and innovations},
  author={Michailidis, Panagiotis and Michailidis, Iakovos and Kosmatopoulos, Elias},
  journal={Energies},
  volume={18},
  number={7},
  pages={1724},
  year={2025},
  publisher={MDPI}
}

@article{cai2021machine,
  title={A machine learning-based model predictive control method for pumped storage systems},
  author={Cai, Qingsen and Luo, Xingqi and Gao, Chunyang and Guo, Pengcheng and Sun, Shuaihui and Yan, Sina and Zhao, Peiyu},
  journal={Frontiers in Energy Research},
  volume={9},
  pages={757507},
  year={2021},
  publisher={Frontiers Media SA}
}

@article{arroyo2022reinforced,
  title={Reinforced model predictive control (RL-MPC) for building energy management},
  author={Arroyo, Javier and Manna, Carlo and Spiessens, Fred and Helsen, Lieve},
  journal={Applied Energy},
  volume={309},
  pages={118346},
  year={2022},
  publisher={Elsevier}
}

@article{rulff2025systematic,
  title={Systematic refinement of surrogate modelling procedure for useful application to building energy problems},
  author={Rulff, David and Evins, Ralph},
  journal={Journal of Building Performance Simulation},
  volume={18},
  number={4},
  pages={389--423},
  year={2025},
  publisher={Taylor \& Francis}
}

@article{ledee2025improved,
  title={Improved surrogate modeling for multi-energy system design: Model architecture, sampling and scaling choices},
  author={L{\'e}d{\'e}e, Fran{\c{c}}ois and Crawford, Curran and Evins, Ralph},
  journal={Applied Energy},
  volume={390},
  pages={125812},
  year={2025},
  publisher={Elsevier}
}

@article{gao2025online,
  title={Online optimization of integrated energy systems based on deep learning predictive control},
  author={Gao, Yuefen and Zhang, Yiying and Yun, Chengbo and Huang, Lizhuang},
  journal={Electric Power Systems Research},
  volume={243},
  pages={111510},
  year={2025},
  publisher={Elsevier}
}

@article{sha2025online,
  title={Online learning-enhanced data-driven model predictive control for optimizing HVAC energy consumption, indoor air quality and thermal comfort},
  author={Sha, Xinyi and Ma, Zhenjun and Sethuvenkatraman, Subbu and Li, Wanqing},
  journal={Applied Energy},
  volume={383},
  pages={125341},
  year={2025},
  publisher={Elsevier}
}

@Online{nrel_eff,
	author  = {NREL},
	title   = {Champion Module Efficiencies},
	url     = {https://www.nrel.gov/docs/libraries/pv/champion-module-efficiencies.pdf},
	urldate = {2025-07-15},
	year    = {2025},
}

@Report{JordanKurtz2013,
	author = {Jordan, D. C. and Kurtz, S. R.},
	title = {Photovoltaic Degradation Rates -- an Analytical Review},
	journal = {Progress in Photovoltaics: Research and Applications},
	volume = {21},
	number = {1},
	pages = {12-29},
	doi = {10.1002/pip.1182},
	url = {https://doi.org/10.1002/pip.1182},
	year = "2013"
}

\end{document}